\RequirePackage{fix-cm} 
\PassOptionsToPackage{table}{xcolor}
\documentclass{article}
\usepackage{iclr2027_conference,times}
\usepackage[T1]{fontenc}
\usepackage[utf8]{inputenc}
\usepackage{amsmath,amssymb,mathtools}
\usepackage{graphicx,booktabs,multirow,tabularx,longtable,array}
\usepackage{float}
\usepackage{wrapfig}
\usepackage{needspace}
\definecolor{paperblue}{RGB}{70,130,180}
\definecolor{paperred}{RGB}{190,90,90}
\AddToHook{env/table/begin}{\setlength{\belowcaptionskip}{6pt}}
\usepackage{tikz}
\usetikzlibrary{arrows.meta}
\usepackage[ruled,vlined]{algorithm2e}
\usepackage{hyperref,url}
\IfFileExists{latexml.sty}{\usepackage{latexml}}{\newif\iflatexml}
\newcommand{\surveyDatasets}{four}
\newcommand{\surveyTasks}{716}
\newcommand{\surveyReschedulable}{266}
\newcommand{\surveyReschedulablePct}{37.2}
\newcommand{\surveyDiverse}{9}
\newcommand{\surveyDiversePct}{3.4}
\newcommand{\surveyLockedPct}{96.6}
\newcommand{\surveyMacroLockedPct}{76.3}

\hypersetup{hidelinks,pdftitle={Learn the Task, Not the Schedule: Reducing Spurious Cross-Arm Coupling in Bimanual Imitation Learning},pdfauthor={Shuyuan Hu, Deng-Ke Wu, Yifan Jin, Qianhang Luo, Mingchen Ju, Kai Ye, Bo Zhao}}
\DeclareRobustCommand{\paperref}[2]{\hyperref[#2]{#1~\getrefnumber{#2}}}
\newcommand{\method}{CTR}
\newcommand{\E}{\mathbb{E}}
\newcommand{\ind}{\mathbb{I}}

\newcommand{\rev}[1]{#1}
\DeclareMathOperator*{\argmin}{arg\,min}
\iclrfinalcopy
\title{Learn the Task, Not the Schedule:\\Reducing Spurious Cross-Arm Coupling\\in Bimanual Imitation Learning}
\iflatexml
\author{%
Shuyuan Hu\footnotemark[1]\\
School of AI, Shanghai Jiao Tong University\\
hushuyuan42@gmail.com
\And
Deng-Ke Wu\footnotemark[1]\\
School of AI, Shanghai Jiao Tong University
\And
Yifan Jin\\
School of AI, Shanghai Jiao Tong University\\
Central South University
\And
Qianhang Luo\\
School of AI, Shanghai Jiao Tong University\\
South China Normal University
\And
Mingchen Ju\\
University of Sydney
\And
Kai Ye\\
School of AI, Shanghai Jiao Tong University
\And
Bo Zhao\footnotemark[2]\\
School of AI, Shanghai Jiao Tong University\\
bo.zhao@sjtu.edu.cn
}
\else
\author{%
Shuyuan Hu\textsuperscript{1,*}\quad
Deng-Ke Wu\textsuperscript{1,*}\quad
Yifan Jin\textsuperscript{1,2}\quad
Qianhang Luo\textsuperscript{1,3}\\
\bfseries Mingchen Ju\textsuperscript{4}\quad
Kai Ye\textsuperscript{1}\quad
Bo Zhao\textsuperscript{1,\ensuremath{\dagger}}\\[4pt]
\normalfont\small \textsuperscript{1}School of AI, Shanghai Jiao Tong University\\
\normalfont\small \textsuperscript{2}Central South University\quad
\textsuperscript{3}South China Normal University\quad
\textsuperscript{4}University of Sydney\\
\normalfont\small \texttt{hushuyuan42@gmail.com}\quad
\texttt{bo.zhao@sjtu.edu.cn}
}
\fi
\begin{document}
\raggedbottom
\maketitle
\lhead{Preprint}
\iflatexml
\footnotetext[1]{Shuyuan Hu and Deng-Ke Wu contributed equally.}
\footnotetext[2]{Corresponding author: Bo Zhao.}
\else
\begingroup
\renewcommand{\thefootnote}{\fnsymbol{footnote}}
\footnotetext[1]{Equal contribution.}
\footnotetext[2]{Corresponding author.}
\endgroup
\fi

\begin{abstract}
{
A bimanual demonstration specifies both \emph{what} the robot should do and one particular schedule for \emph{when} each arm should act.
Standard imitation learning does not distinguish task-required cross-arm constraints from incidental scheduling choices, and may therefore exploit opposite-arm progress as a shortcut cue, creating \emph{spurious cross-arm coupling} even where relative timing is irrelevant for the current task phase.
We introduce Constraint-preserving Temporal Recombination (\method), a training-time data intervention that preserves recorded arm-local sensorimotor content while retiming the corresponding segments to realize admissible schedules that preserve task-required cross-arm constraints.
We use \method{} to train Diffusion Policy, Action Chunking with Transformers, and $\pi_{0.5}$, and evaluate them on four simulated and four real-robot tasks.
Across six real-robot bimanual progress-state initializations, \method{} raises worst-state success from 56\% for the strongest non-CTR schedule mixture to 76\%.
With $\pi_{0.5}$ on an Independent task, a 4\,s pause transfers 2.56\,s of delay to the opposite arm under the Concurrent schedule distribution, but only 0.25\,s with \method{}.
On real robots, \method{} violates task constraints in only 0--4\% of trials, versus 36--98\% for narrow-schedule baselines.
Code will be released at \url{https://github.com/MINT-SJTU/CTR}.
\looseness=-1\par
}
\end{abstract}


\section{Introduction}
{
In a bimanual demonstration, the two arms follow one of possibly many admissible schedules. If each arm independently picks an object and the demonstrator does so sequentially, the recorded order is a choice, not a task requirement; likewise, a drawer must be opened before an object is inserted, but not before the object is grasped. Expressive imitation policies such as Diffusion Policy (DP), Action Chunking with Transformers (ACT), and vision-language-action models including $\pi_0$, $\pi_{0.5}$, and OpenVLA~\citep{chi2023diffusion,zhao2023act,black2025pi0,pi2025pi05,kim2025openvla} cannot tell these cases apart from the demonstrations alone. When demonstrations cover only a few schedules, one arm's behavior can come to depend on the other's progress even where their subtasks are independently schedulable; we call this \emph{spurious cross-arm coupling}.
\looseness=-1\par
}

The narrow schedule support that creates this ambiguity appears common in existing datasets. Our audit covers \surveyDatasets{} public bimanual datasets: RoboTwin~2.0, RoboCOIN, DexMimicGen, and BiCoord~\citep{chen2025robotwin2,wu2025robocoin,jiang2025dexmimicgen,peng2026bicoord}. It finds that \surveyReschedulable{} of \surveyTasks{} tasks (\surveyReschedulablePct{}\%) are reschedulable, yet only \surveyDiversePct{}\% of these exhibit multiple observed admissible schedules (\paperref{Appendix}{app:dataset_survey}).
{
This limited coverage motivates learning cross-arm dependence that reflects task constraints rather than the particular schedules demonstrated. We call this \emph{selective cross-arm dependence}: in independently schedulable phases, delaying one arm should not make the other wait, whereas during insertion, support, or shared-workspace occupancy, the policy must respond to the opposite arm. More demonstrations with the same narrow schedule support cannot resolve this ambiguity (\paperref{Section}{sec:identifiability}).
\looseness=-1\par
}

We propose \textbf{Constraint-preserving Temporal Recombination (\method{})}, a training-time data intervention that broadens schedule support without new demonstrations. \method{} segments each arm's recorded stream and retimes the segments into alternative admissible schedules while preserving task-required cross-arm constraints, pairing the same arm-local sensorimotor content with varied opposite-arm progress.

{
We evaluate \method{} with three policy families (DP, ACT, and $\pi_{0.5}$) on four simulated and four real-robot tasks. \method{}-trained policies achieve higher default-condition success rates than the same architectures trained with serial schedules in 19 of 24 model--task comparisons, and \method{} raises real-robot worst-case progress-state success from 56\% to 76\%. Pausing one arm for 4\,s delays the other arm's completion by 2.56\,s with $\pi_{0.5}$ trained on concurrent-start schedules, but by only 0.25\,s with \method{}. Required coordination is retained: under real-robot constraint interventions, \method{}-trained $\pi_{0.5}$ violates task constraints in 0\% and 4\% of trials, versus 98\% and 36\% for narrow-schedule baselines.
\looseness=-1\par
}

{
Our contributions are: (1) identifying spurious cross-arm coupling under narrow schedule support and formalizing selective cross-arm dependence as the desired behavior; (2) \method{}, which broadens admissible schedule support while preserving arm-local sensorimotor content and task-required cross-arm constraints; and (3) simulated and real-robot evidence across three policy families that \method{} reduces spurious coupling without sacrificing required coordination.
\looseness=-1\par
}

\section{Related work}
\paragraph{Bimanual policies and cross-arm dependence.}
{
Bimanual imitation policies model cross-arm dependence through joint action chunks in ACT~\citep{zhao2023act} or explicit attention and synchronization in InterACT~\citep{lee2025interact}.
BUDS structures this dependence through acting and stabilizing roles~\citep{grannen2023stabilize}.
To support independence alongside coordination, the Decoupled Interaction Framework combines arm-specific policies with selective information exchange~\citep{jiang2025dif}, while SkillVLA reuses single-arm skills across novel left--right pairings~\citep{zhai2026skillvla}.
These approaches structure dependence at the model or skill level; our focus is the training distribution, where narrow schedule support can make opposite-arm progress predictive even when it is irrelevant to the current arm-local subtask.
\looseness=-1\par
}

\vspace{-0.5ex}
\paragraph{Causal confusion and counterfactual augmentation.}
{
This mismatch between predictiveness and task relevance connects schedule-induced coupling to causal confusion in imitation learning~\citep{dehaan2019causal}.
A related data-level strategy is counterfactual augmentation: CoDA recombines locally factored transitions~\citep{pitis2020coda}, and CAIAC exchanges state components estimated to be unaffected by an action~\citep{urpi2024caiac}.
RoCoDA brings this principle to visual robot imitation, including bimanual tasks, by recombining state partitions under phase-specific causal structure~\citep{ameperosa2025rocoda}.
Here, the intervention of interest is relative arm timing while the recorded arm-local behavior remains fixed.
Because relative timing can also encode required coordination, this intervention must vary progress pairings without violating task-required cross-arm constraints.
\looseness=-1\par
}

\vspace{-0.5ex}
\paragraph{Constraint-aware generation and temporal recombination.}
{
Demonstration generation provides tools for this constrained recombination: MimicGen adapts object-centric segments to new contexts~\citep{mandlekar2023mimicgen}, and DexMimicGen adds asynchronous per-arm execution with synchronization and ordering constraints~\citep{jiang2025dexmimicgen}.
TriManPolicy more directly addresses interface-induced waiting by retiming fixed-duration sensorimotor segments under dependency and arm-resource constraints to minimize demonstration makespan~\citep{zhao2026triman}.
However, selecting an efficient schedule does not by itself establish robustness to other admissible schedules.
\method{} therefore samples alternative schedules for the same recorded arm-local sensorimotor content under annotated task constraints, broadening progress-state support without changing the policy architecture or deployment procedure.
To distinguish selective cross-arm dependence from indiscriminate decoupling, we combine held-out progress tests and paired opposite-arm pauses with task-constraint interventions.
\looseness=-1\par
}

\section{Problem formulation: spurious coupling versus selective dependence}
\subsection{\rev{Bimanual imitation learning}}
{
\rev{We learn a joint bimanual policy from demonstrations $\mathcal D=\{\tau_n\}_{n=1}^N$ with observations $o_t$ and joint actions $a_t=(a_t^L,a_t^R)$. For a base learner $m$, let $h_t$ contain the current observation, available history, and task instruction when provided; $\pi_{\theta_m}(A_t\mid h_t)$ predicts an $H_m$-step action chunk $A_t=(a_t,\ldots,a_{t+H_m-1})$. Let $P_{\mathcal D}$ be the empirical trajectory distribution and $\mathcal W_m(\tau)$ the learner's distribution over valid training windows $w=(h_t,A_t)$ from $\tau$. We minimize the original training loss:}
\looseness=-1\par
}
\begin{equation}
 \theta_{m,\rm BC}^\star
 =\argmin_{\theta_m}\;
 \E_{\tau\sim P_{\mathcal D}}\,
 \E_{w\sim\mathcal W_m(\tau)}
 \bigl[\mathcal L_m(\theta_m;w)\bigr].
 \label{eq:bc}
\end{equation}
\rev{Here $\mathcal L_m$ is the base learner's per-window loss, including model-specific action and auxiliary terms; the expectations average it over empirical training windows. These windows pair both arms on the recorded timeline, so the training data also encode the demonstrator schedule.}

\subsection{\rev{Spurious cross-arm coupling}}
{
\rev{A demonstration records one demonstrator schedule although the task may permit other admissible schedules with the same arm-local sensorimotor content. We call this limited support over admissible schedules \emph{narrow schedule support}. Under narrow schedule support, imitation may treat opposite-arm progress as a shortcut, producing \emph{spurious cross-arm coupling} when the policy responds to an incidental timing relation.}
\looseness=-1\par
}

\rev{For fixed arm-local sensorimotor content in $\tau$, let $\mathcal S(\tau)$ contain schedules respecting within-arm order, $V_{\rm task}(\tau,s)$ encode physical and semantic validity, and $\mathcal R(\tau)$ contain candidate cross-arm relations observed in $\tau$. Define}
\begin{align}
 \Omega^\star(\tau)
 &=\{s\in\mathcal S(\tau):V_{\rm task}(\tau,s)=1\},
 \label{eq:admissible-schedules}\\
 \mathcal R_{\rm req}^\star(\tau)
 &=\{r\in\mathcal R(\tau):r(s)=1\ \forall s\in\Omega^\star(\tau)\}.
 \label{eq:required-constraints}
\end{align}
\rev{Relations in $\mathcal R_{\rm req}^\star(\tau)$ are \emph{task-required cross-arm constraints}; a candidate relation that changes under some admissible schedule is \emph{reschedulable}. Relations specify properties such as order or non-overlap, not exact delays: a required order can allow timing slack. This is a property of the task, not the learned policy.}

\paragraph{\rev{Selective cross-arm dependence.}}
\rev{Let $x_i$ be arm $i$'s local task state, $p_i$ its within-segment progress, and $c$ the shared conditions relevant to coordination. We call $z=(p_L,p_R)$ the \emph{bimanual progress state}. In an independently schedulable phase, the arm-$i$ marginal should be approximately invariant to an admissible change in opposite-arm progress $p_j$ ($j\neq i$):}
\begin{equation}
 \pi_{\theta_m}^i(\cdot\mid x_i,p_i,c,p_j)
 \approx \pi_{\theta_m}^i(\cdot\mid x_i,p_i,c,p'_j).
 \label{eq:invariance}
\end{equation}
\rev{In a coupled phase, shared task progress $p_C$ cannot be split into independently schedulable arm-local progress; a change in a prerequisite, support, or workspace condition can instead require a different response. We test both sides through phase-conditioned sensitivity and task-constraint interventions.}

\subsection{What is and is not identifiable}
\label{sec:identifiability}
{
\rev{If every demonstration places one arm's segment before the other's, that order may be a prerequisite or a habit. Under an idealized fixed schedule, $p_j=h(x_i,p_i,c)$ almost surely; dependent and independent policies can therefore agree on observed bimanual progress states but differ at unobserved admissible states. More demonstrations drawn from the same narrow schedule support cannot distinguish them.}
\looseness=-1\par
}

\rev{Determining which alternatives are safe requires task knowledge beyond the observed timing. CTR supplies it through task-constraint annotations during data construction; the deployed policy does not receive these annotations.}

\section{Constraint-Preserving Temporal Recombination}
\rev{Recollecting demonstrations under different schedules could also change the arm-local trajectories. CTR instead preserves recorded arm-local sensorimotor content while retiming segments to realize alternative admissible schedules, making the schedule the intended intervention and broadening schedule support without new demonstrations (\paperref{Figure}{fig:method}).}

\subsection{Task-specific automatic segmentation and constraint extraction}
\label{sec:annotate}
{
\rev{Let $\tau=\{(o_t,a_t)\}_{t=1}^{T}$ be a recorded demonstration with per-arm streams $\tau^i$, $i\in\{L,R\}$. At task setup time, GPT-6 Astra~\citep{openai2026astra} generates a task-specific video-processing pipeline $\mathcal P_\xi$. The generated program detects task-relevant interactions from the available visual and robot evidence, decomposes each arm-local stream into $K_i$ ordered segments $\sigma_{i,1}\prec\cdots\prec\sigma_{i,K_i}$ tiling the demonstration timeline, $\sigma_{i,k}=[u_{i,k},v_{i,k}]$ with duration $\ell_{i,k}=v_{i,k}-u_{i,k}$, and emits the task-required cross-arm relations used for schedule construction and validation. We write $\mathcal R_{\rm req}(\tau)$ for these pipeline-produced constraints on segment pairs; they approximate $\mathcal R_{\rm req}^\star(\tau)$ rather than being inferred from the recorded schedule alone.}
\looseness=-1\par
}

{
\rev{A schedule $s=\{s_{i,k}\}$ places each segment at $\Sigma_{i,k}(s)=[s_{i,k},\,s_{i,k}+\ell_{i,k}]$, preserving its content and duration. The implemented set $\Omega(\tau)$ keeps within-arm order and freezes the recorded relative offsets of segment pairs governed by constraints in $\mathcal R_{\rm req}(\tau)$. Intended as a conservative subset of $\Omega^\star(\tau)$, it may exclude admissible timing variations under those constraints while allowing reschedulable blocks to become serial or overlapping.}
\looseness=-1\par
}

\subsection{Retime segments while preserving arm-local sensorimotor content}
\label{sec:recombine}
\rev{For $s\in\Omega(\tau)$, the source-time map $\phi_i^s$ advances through arm $i$'s active segments and holds during inserted gaps (\paperref{Equation}{eq:phi}, \paperref{Appendix}{app:implementation}). The recombination $\widetilde\tau=\mathcal C(\tau,\phi_L^s,\phi_R^s)$ reads arm-local streams at their respective source times and reconstructs the shared main view (\paperref{Section}{sec:realdata}). For every active segment and offset $u\in[0,\ell_{i,k}]$, it preserves recorded arm-local sensorimotor content:}
\begin{equation}
 (\widetilde o^{\,i}_{s_{i,k}+u},\widetilde a^{\,i}_{s_{i,k}+u})
 =(o^i_{u_{i,k}+u},a^i_{u_{i,k}+u}).
 \label{eq:preserve}
\end{equation}

\rev{CTR keeps the empirical demonstration law $\tau\sim P_{\mathcal D}$ but replaces the recorded schedule law $\delta_{s_\tau}$ with $s\sim q(\cdot\mid\tau)$ supported on $\Omega(\tau)$. It forms $\widetilde\tau=\mathcal C(\tau,\phi_L^s,\phi_R^s)$ and extracts training windows. The implementation samples admissible reschedulable-block offsets uniformly, pairing the same arm-local sensorimotor content with varied opposite-arm progress.}

\rev{Let $M_s$ mask only action-supervision coordinates belonging to synthetic waits; model-specific auxiliary terms remain outside the mask normalization. We minimize the CTR training loss:}
\begin{equation}
 \theta_{m,\rm CTR}^\star
 =\argmin_{\theta_m}\;
 \E_{\tau\sim P_{\mathcal D}}\,
 \E_{s\sim q(\cdot\mid\tau)}\,
 \E_{w\sim\mathcal W_m(\mathcal C(\tau,\phi_L^s,\phi_R^s))}
 \bigl[\mathcal L_m(\theta_m;w,M_s)\bigr],
 \label{eq:objective}
\end{equation}
\rev{\paperref{Equation}{eq:loss} specifies the masked action term. If $q(\cdot\mid\tau)=\delta_{s_\tau}$, $\mathcal C(\tau,\phi_L^{s_\tau},\phi_R^{s_\tau})=\tau$, and $M_{s_\tau}=1$, \paperref{Equation}{eq:objective} reduces to the original training objective in \paperref{Equation}{eq:bc}.}

\begin{figure}[htbp]
\centering
\includegraphics[width=\linewidth]{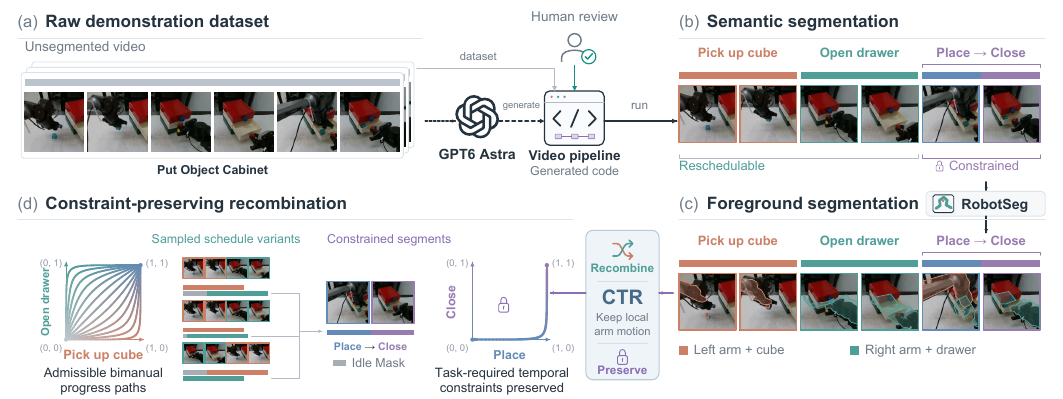}
\caption{\method{} data-processing pipeline. Foreground segmentation supports observation composition, and temporal recombination realizes alternative admissible schedules while preserving arm-local sensorimotor content and task-required cross-arm constraints. The displayed masks and sampled schedules are schematic.}
\label{fig:method}
\end{figure}

\subsection{CTR on recorded real-robot data: two implementation issues}
\label{sec:realdata}
{
\rev{For the main camera, foreground segmentation extracts the two arms and task-relevant objects, and we composite them at the recombined source times $\phi_i^s(t)$ over the scene background, approximating both arms executing the recombined schedule concurrently (\paperref{Equation}{eq:compose}, \paperref{Appendix}{app:implementation}). For supervision, an IdleMask zeroes the imitation loss on coordinates and timesteps whose waiting was inserted by recombination, while required holding, active support, and terminal holding remain supervised (\paperref{Equations}{eq:idlemask}--\hyperref[eq:loss]{\getrefnumber{eq:loss}}, \paperref{Appendix}{app:implementation}). Preserving arm-local streams does not by itself guarantee that every recombined joint scene is physically executable, nor that a composed observation is identical to a concurrently captured real observation. We therefore apply task-specific geometric, contact, and observation-consistency checks as conservative filters; \paperref{Appendix}{app:physical-validity} details these audits and their remaining limitations.}
\looseness=-1\par
}

\section{Experiments}
\label{sec:experiments}
We first diagnose spurious cross-arm coupling at the action and rollout levels, then test whether CTR reduces it while preserving task-required dependence, and finally ask whether the resulting robustness extends beyond training support.

\subsection{Experimental setup}
\label{sec:exp-setup}
\paragraph{Tasks, models, and evaluation.}
{
We evaluate DP, ACT, and $\pi_{0.5}$ on four simulated and four real-robot tasks spanning \textbf{Independent}, \textbf{Shared-Workspace Exclusion}, \textbf{Coordination}, and \textbf{Prerequisite} relations (\paperref{Figure}{fig:task-overview}). The simulated tasks are Sort Blocks, Place Dual Shoes, Scan Object, and Put Object Cabinet. Sort Blocks is slightly modified from RoboTwin~2.0's Blocks Ranking task~\citep{chen2025robotwin2}, while the other three are taken directly from RoboTwin~2.0. The real-robot tasks are Sort Letters, Organize Hardware, Deliver Water, and Put Object Cabinet. Task details and modifications are provided in \paperref{Appendix}{app:tasks}. \paperref{Appendix}{app:training-evaluation-config} gives the training and simulation-evaluation configurations, and \paperref{Appendix}{app:metrics} defines the metrics and aggregation conventions. All success rates reported in this paper are computed from 50 evaluation trials.
\looseness=-1\par
}

\begingroup
\setlength{\intextsep}{6pt}
\begin{figure}[htbp]
\centering
\makebox[\linewidth][c]{\includegraphics[width=1.08\linewidth,trim=0 18bp 0 25bp,clip]{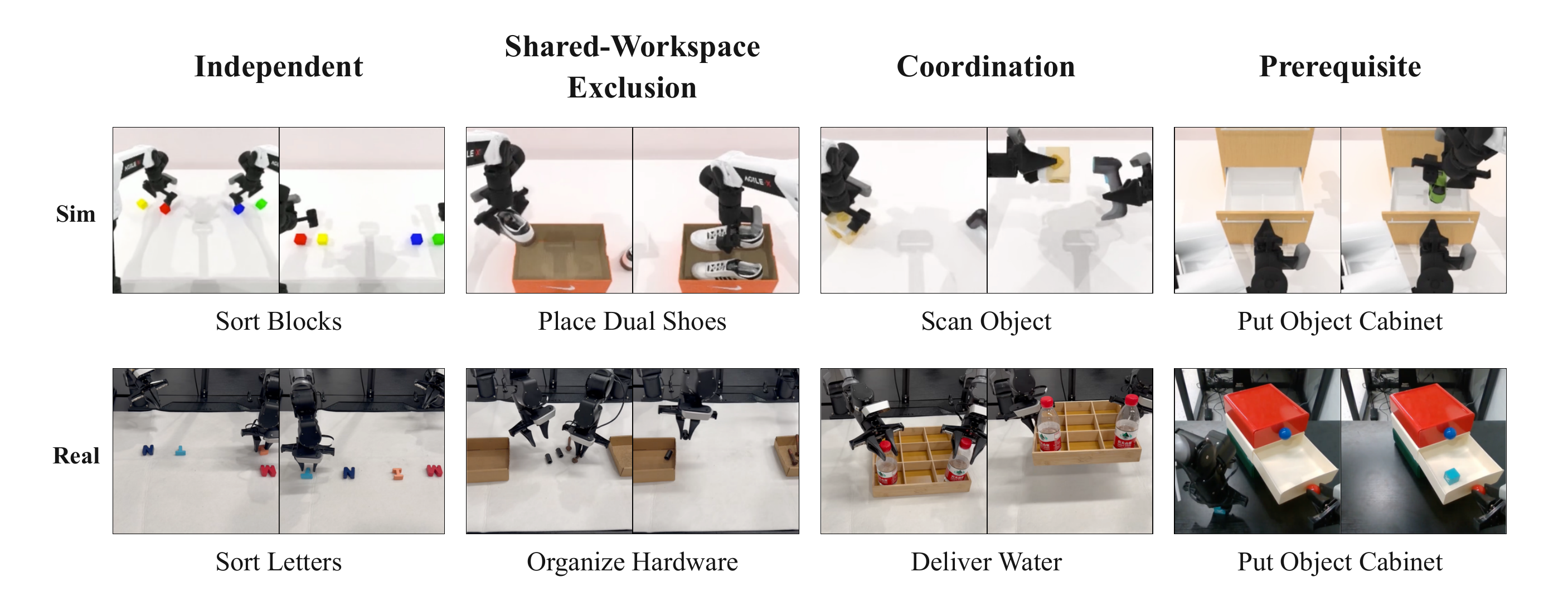}}%
\caption{The eight evaluation tasks span Independent, Shared-Workspace Exclusion, Coordination, and Prerequisite relations, testing both invariance where relative timing is irrelevant and responsiveness where cross-arm constraints matter. Columns group tasks by relation type; the top and bottom rows show simulated and real-robot tasks, respectively, with two representative frames per task.}
\label{fig:task-overview}
\end{figure}
\endgroup

\paragraph{Training schedule distributions.}
{
To vary schedule support while preserving arm-local sensorimotor content and annotated task-required constraints, we compare five distributions over independently schedulable regions: $\boldsymbol{L\!\rightarrow\!R}$ (one serial order), \textbf{Serial} (both serial orders), \textbf{Concurrent}, \textbf{Mixed} (both serial orders plus Concurrent), and \textbf{\method} (relative timing sampled across the admissible range). Serial, Concurrent, Mixed, and CTR share the main-camera visual-synthesis pipeline in \paperref{Section}{sec:realdata}; the original-data $L\!\rightarrow\!R$ condition retains recorded main-camera observations. Only CTR uses IdleMask, because its recombination can insert artificial waits. \paperref{Section}{sec:ablation} controls for both visual synthesis and action-supervision masking. \paperref{Figure}{fig:schedule-distributions} shows empirical onset-difference densities, and \paperref{Appendices}{app:schedule-distributions} and~\hyperref[app:implementation]{\getrefnumber{app:implementation}} detail distribution construction and implementation.
\looseness=-1\par
}

\begin{figure}[t]
\centering
\input{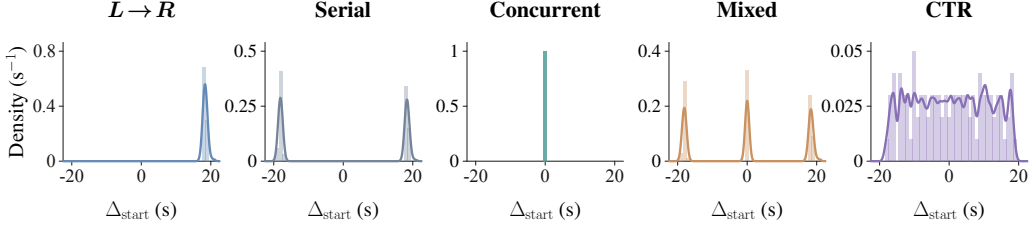}
\caption{The five training distributions vary relative arm timing: the baselines concentrate on one or more canonical schedules, whereas CTR covers a broader range of admissible timings. Histograms show onset-difference densities from Sort Blocks simulation trajectories, with $\Delta_{\mathrm{start}}=t_R^{\rm start}-t_L^{\rm start}>0$ indicating that the left arm starts first; vertical density scales differ across panels. Curves show smoothed density estimates except for Concurrent; smoothing-induced tails should not be interpreted as additional observed support. \paperref{Appendix}{app:schedule-distributions} gives construction and measurement details.}
\label{fig:schedule-distributions}
\end{figure}

\subsection{Diagnosing spurious cross-arm coupling}
\label{sec:exp-diagnosis}
\paragraph{Action-level sensitivity in independently schedulable phases.}
{
The invariance in \paperref{Equation}{eq:invariance} motivates a phase-conditioned action-sensitivity probe (\paperref{Appendix}{app:phase-sensitivity-probe}) on Deliver Water, whose independent \emph{Place Bottles} and \emph{Return to Standby} phases surround the synchronized \emph{Serve Water} phase. \paperref{Figure}{fig:phase-conditioned-sensitivity} compares mean sensitivity curves for ACT, DP, and $\pi_{0.5}$. Averaging matched interventions in both directions, Concurrent ACT yields $(S_{\rm place},S_{\rm serve},S_{\rm return})=(13.28,\,5.82,\,9.77)$, with greater sensitivity in both independent phases than during synchronization; CTR ACT yields $(2.79,\,4.19,\,1.07)$, suppressing independent-phase sensitivity while retaining a \emph{Serve Water} response larger than its own independent-phase responses. CTR therefore reduces sensitivity selectively in the independent phases while retaining a stronger response during synchronization.
\looseness=-1\par
}

\begin{figure*}[t]
\centering
\includegraphics[width=\linewidth]{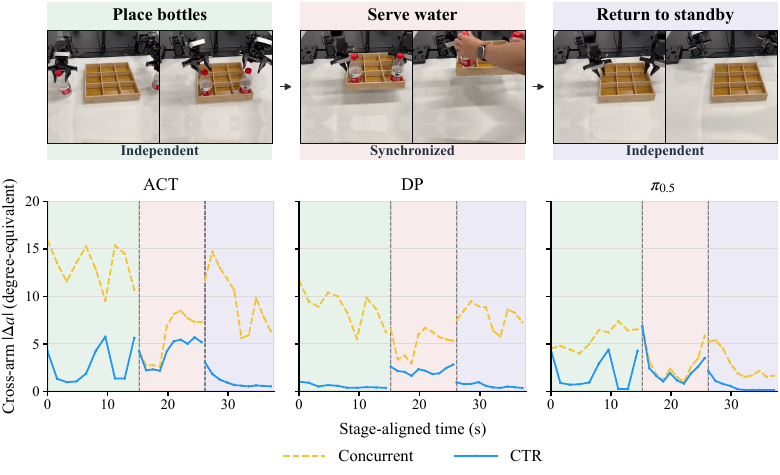}
\caption{Phase-conditioned action sensitivity on Deliver Water for ACT, DP, and $\pi_{0.5}$. We measure changes in one arm's predicted actions while keeping its state and observations fixed and varying the opposite arm's state and observations. Curves show the mean action difference for samples within each time interval. CTR reduces sensitivity during the independent phases while retaining responsiveness during \emph{Serve Water}. \paperref{Appendix}{app:phase-sensitivity-probe} shows the original scatter plots for all three models.}
\label{fig:phase-conditioned-sensitivity}
\end{figure*}

\paragraph{From action dependence to progress-state brittleness.}
{
If a policy uses opposite-arm progress as a shortcut, a familiar local state paired with a different admissible opposite-arm state can elicit an inappropriate action and compromise task completion. We test this behavioral prediction with $\pi_{0.5}$ on Put Object Cabinet in simulation and Sort Letters on the real robot. Each task uses six task-defined initial states in independently schedulable regions: nominal (C00), matched-progress (Cmid), left-leading, right-leading, and two task-specific ready/completion-side states (CL/CR); these are not interpolated robot joint configurations. \paperref{Table}{tab:progress_robustness} reports success rate (SR) and failure-penalized completion time $T_{\rm Pen}$. On Sort Letters, $L\!\rightarrow\!R$, Serial, and Concurrent achieve 90--92\% SR from C00, yet their worst-state SRs fall to 36\%, 0\%, and 28\%, respectively. On Put Object Cabinet, Concurrent reaches 98\% at Cmid but 0\% at all four asymmetric states. Narrow serial and concurrent support can thus both leave policies brittle across admissible bimanual progress states. These rollout results extend the action-level probe to closed-loop behavior across tasks and models.
\looseness=-1\par
}

\begin{table}[t]
\caption{Bimanual progress-state robustness for $\pi_{0.5}$ on the Put Object Cabinet task from the RoboTwin~2.0 benchmark~\citep{chen2025robotwin2} and on Sort Letters on the real robot. During phases in which the two arms' subtasks are independent, we expect policy performance to remain unchanged across initializations with different relative progress between the arms. Each measured-state cell ($n=50$) reports SR (\%) / $T_{\rm Pen}$ (s). When SR=0, $T_{\rm Pen}$ is undefined and shown as ``--'' because its failure penalty depends on the cell's mean successful completion time. Cell tint is determined solely by SR, with darker blue indicating higher success rates. Worst SR is the minimum over the six states. \paperref{Appendix}{app:metrics} specifies the penalty.}
\label{tab:progress_robustness}
\centering\setlength{\tabcolsep}{1.5pt}
\AddToHookNext{env/tabular/begin}{\color{black}}
\begin{tabular}{llcccccc|c}
\toprule
Task & Data & C00 & Cmid & L-lead & R-lead & CL & CR & Worst SR \\
\midrule
\multirow{5}{*}{\parbox{1.5cm}{\raggedright Put Object Cabinet (sim)}} & $L\!\rightarrow\!R$ & \cellcolor{paperblue!14}\makebox[1.65em][r]{40}\,/\,\makebox[1.75em][l]{16.4} & \cellcolor{paperblue!19.6}\makebox[1.65em][r]{56}\,/\,\makebox[1.75em][l]{3.8} & \cellcolor{paperblue!25.9}\makebox[1.65em][r]{74}\,/\,\makebox[1.75em][l]{6.4} & \makebox[1.65em][r]{0}\,/\,\makebox[1.75em][l]{--} & \cellcolor{paperblue!21}\makebox[1.65em][r]{\textbf{60}}\,/\,\makebox[1.75em][l]{9.1} & \cellcolor{paperblue!0.7}\makebox[1.65em][r]{2}\,/\,\makebox[1.75em][l]{8.1} & 0 \\
 & Serial & \cellcolor{paperblue!12.6}\makebox[1.65em][r]{36}\,/\,\makebox[1.75em][l]{16.9} & \cellcolor{paperblue!31.5}\makebox[1.65em][r]{90}\,/\,\makebox[1.75em][l]{3.6} & \cellcolor{paperblue!26.6}\makebox[1.65em][r]{\textbf{76}}\,/\,\makebox[1.75em][l]{6.4} & \cellcolor{paperblue!10.5}\makebox[1.65em][r]{30}\,/\,\makebox[1.75em][l]{8.8} & \cellcolor{paperblue!18.2}\makebox[1.65em][r]{52}\,/\,\makebox[1.75em][l]{9.3} & \cellcolor{paperblue!15.4}\makebox[1.65em][r]{44}\,/\,\makebox[1.75em][l]{10.6} & \cellcolor{paperblue!10.5}30 \\
 & Concurrent & \cellcolor{paperblue!7.7}\makebox[1.65em][r]{22}\,/\,\makebox[1.75em][l]{12.1} & \cellcolor{paperblue!34.3}\makebox[1.65em][r]{98}\,/\,\makebox[1.75em][l]{2.9} & \makebox[1.65em][r]{0}\,/\,\makebox[1.75em][l]{--} & \makebox[1.65em][r]{0}\,/\,\makebox[1.75em][l]{--} & \makebox[1.65em][r]{0}\,/\,\makebox[1.75em][l]{--} & \makebox[1.65em][r]{0}\,/\,\makebox[1.75em][l]{--} & 0 \\
 & Mixed & \cellcolor{paperblue!18.9}\makebox[1.65em][r]{54}\,/\,\makebox[1.75em][l]{10.1} & \cellcolor{paperblue!35}\makebox[1.65em][r]{\textbf{100}}\,/\,\makebox[1.75em][l]{3.0} & \cellcolor{paperblue!22.4}\makebox[1.65em][r]{64}\,/\,\makebox[1.75em][l]{7.0} & \cellcolor{paperblue!17.5}\makebox[1.65em][r]{50}\,/\,\makebox[1.75em][l]{7.4} & \cellcolor{paperblue!16.8}\makebox[1.65em][r]{48}\,/\,\makebox[1.75em][l]{10.1} & \cellcolor{paperblue!28}\makebox[1.65em][r]{80}\,/\,\makebox[1.75em][l]{7.8} & \cellcolor{paperblue!16.8}48 \\
 & CTR (ours) & \cellcolor{paperblue!23.1}\makebox[1.65em][r]{\textbf{66}}\,/\,\makebox[1.75em][l]{11.0} & \cellcolor{paperblue!35}\makebox[1.65em][r]{\textbf{100}}\,/\,\makebox[1.75em][l]{3.2} & \cellcolor{paperblue!25.2}\makebox[1.65em][r]{72}\,/\,\makebox[1.75em][l]{6.1} & \cellcolor{paperblue!21.7}\makebox[1.65em][r]{\textbf{62}}\,/\,\makebox[1.75em][l]{11.3} & \cellcolor{paperblue!19.6}\makebox[1.65em][r]{56}\,/\,\makebox[1.75em][l]{8.7} & \cellcolor{paperblue!33.6}\makebox[1.65em][r]{\textbf{96}}\,/\,\makebox[1.75em][l]{6.6} & \cellcolor{paperblue!19.6}\textbf{56} \\
\midrule
\multirow{5}{*}{\parbox{1.5cm}{\raggedright Sort Letters (real)}} & $L\!\rightarrow\!R$ & \cellcolor{paperblue!31.5}\makebox[1.65em][r]{90}\,/\,\makebox[1.75em][l]{32.1} & \cellcolor{paperblue!28.7}\makebox[1.65em][r]{82}\,/\,\makebox[1.75em][l]{29.1} & \cellcolor{paperblue!12.6}\makebox[1.65em][r]{36}\,/\,\makebox[1.75em][l]{48.7} & \cellcolor{paperblue!25.2}\makebox[1.65em][r]{72}\,/\,\makebox[1.75em][l]{39.3} & \cellcolor{paperblue!32.2}\makebox[1.65em][r]{92}\,/\,\makebox[1.75em][l]{22.2} & \cellcolor{paperblue!34.3}\makebox[1.65em][r]{98}\,/\,\makebox[1.75em][l]{18.0} & \cellcolor{paperblue!12.6}36 \\
 & Serial & \cellcolor{paperblue!32.2}\makebox[1.65em][r]{92}\,/\,\makebox[1.75em][l]{32.7} & \makebox[1.65em][r]{0}\,/\,\makebox[1.75em][l]{--} & \cellcolor{paperblue!6.3}\makebox[1.65em][r]{18}\,/\,\makebox[1.75em][l]{54.6} & \cellcolor{paperblue!17.5}\makebox[1.65em][r]{50}\,/\,\makebox[1.75em][l]{46.8} & \cellcolor{paperblue!33.6}\makebox[1.65em][r]{96}\,/\,\makebox[1.75em][l]{21.0} & \cellcolor{paperblue!30.8}\makebox[1.65em][r]{88}\,/\,\makebox[1.75em][l]{25.8} & 0 \\
 & Concurrent & \cellcolor{paperblue!32.2}\makebox[1.65em][r]{92}\,/\,\makebox[1.75em][l]{19.2} & \cellcolor{paperblue!32.2}\makebox[1.65em][r]{92}\,/\,\makebox[1.75em][l]{12.6} & \cellcolor{paperblue!18.9}\makebox[1.65em][r]{54}\,/\,\makebox[1.75em][l]{34.9} & \cellcolor{paperblue!18.2}\makebox[1.65em][r]{52}\,/\,\makebox[1.75em][l]{35.9} & \cellcolor{paperblue!14.7}\makebox[1.65em][r]{42}\,/\,\makebox[1.75em][l]{42.4} & \cellcolor{paperblue!9.8}\makebox[1.65em][r]{28}\,/\,\makebox[1.75em][l]{48.6} & \cellcolor{paperblue!9.8}28 \\
 & Mixed & \cellcolor{paperblue!25.9}\makebox[1.65em][r]{74}\,/\,\makebox[1.75em][l]{39.0} & \cellcolor{paperblue!32.9}\makebox[1.65em][r]{\textbf{94}}\,/\,\makebox[1.75em][l]{13.5} & \cellcolor{paperblue!20.3}\makebox[1.65em][r]{58}\,/\,\makebox[1.75em][l]{35.6} & \cellcolor{paperblue!19.6}\makebox[1.65em][r]{56}\,/\,\makebox[1.75em][l]{34.3} & \cellcolor{paperblue!33.6}\makebox[1.65em][r]{96}\,/\,\makebox[1.75em][l]{20.5} & \cellcolor{paperblue!25.9}\makebox[1.65em][r]{74}\,/\,\makebox[1.75em][l]{29.3} & \cellcolor{paperblue!19.6}56 \\
 & CTR (ours) & \cellcolor{paperblue!32.9}\makebox[1.65em][r]{\textbf{94}}\,/\,\makebox[1.75em][l]{21.1} & \cellcolor{paperblue!32.9}\makebox[1.65em][r]{\textbf{94}}\,/\,\makebox[1.75em][l]{17.4} & \cellcolor{paperblue!26.6}\makebox[1.65em][r]{\textbf{76}}\,/\,\makebox[1.75em][l]{26.6} & \cellcolor{paperblue!33.6}\makebox[1.65em][r]{\textbf{96}}\,/\,\makebox[1.75em][l]{18.9} & \cellcolor{paperblue!35}\makebox[1.65em][r]{\textbf{100}}\,/\,\makebox[1.75em][l]{16.4} & \cellcolor{paperblue!35}\makebox[1.65em][r]{\textbf{100}}\,/\,\makebox[1.75em][l]{16.2} & \cellcolor{paperblue!26.6}\textbf{76} \\
\bottomrule
\end{tabular}

\vspace{-4pt}
\end{table}

\subsection{CTR restores robustness with selective cross-arm dependence}
\label{sec:exp-ctr}
\label{sec:ablation}
\paragraph{Recovery under broader schedule support.}
{
Because brittleness appears under multiple narrow schedules, we next broaden their support rather than select another fixed order. In the same Sort Letters stress test, Mixed raises worst-state SR to 56\%, and CTR raises it to 76\% (\paperref{Table}{tab:progress_robustness}). Relative to Mixed, CTR improves R-lead from 56\% to 96\% and CR from 74\% to 100\%, while Cmid remains at 94\%. On Put Object Cabinet, worst-state SR similarly increases from 48\% with Mixed to 56\% with CTR. Broader admissible support thus improves worst-case robustness over a mixture limited to canonical schedules.
\looseness=-1\par
}

\paragraph{Implementation controls.}
{
Since CTR also introduces main-camera visual synthesis and action-supervision masking for synthetic waits, we test these implementation choices separately (\paperref{Table}{tab:ablation}). Applying CTR's main-camera segmentation and compositing to $L\!\rightarrow\!R$ without retiming leaves Sort Letters SR at 90\% and changes Scan Object SR from 68\% to 62\%, so visual synthesis alone does not explain an improvement on these tasks. Removing only IdleMask from CTR reduces SR from 94\% to 92\% and from 70\% to 56\%, respectively, showing that action-supervision masking for synthetic waits also contributes, particularly on Scan Object.
\looseness=-1\par
}

\begin{table}[t]
\centering
\begin{minipage}[t]{0.54\linewidth}
\vspace{0pt}
\caption{Implementation-control SRs (\%; $n=50$) on real-robot Sort Letters and simulated Scan Object, testing masking and visual synthesis.}
\label{tab:ablation}
\centering\setlength{\tabcolsep}{3pt}
\begin{tabular}{lcc}
\toprule
Training variant & Sort Letters & Scan Object \\
\midrule
\method{} & \textbf{94} & \textbf{70} \\
\method{} w/o IdleMask & 92 & 56 \\
$L\!\rightarrow\!R$ & 90 & 68 \\
$L\!\rightarrow\!R$ + visual synthesis & 90 & 62 \\
\bottomrule
\end{tabular}

\end{minipage}\hspace{0.05\linewidth}%
\begin{minipage}[t]{0.40\linewidth}
\vspace{0pt}
\setcounter{table}{4}
\caption{Sort Letters SRs (\%; $n=50$) for $\pi_{0.5}$ at initial states and in a region withheld from CTR-hole training.}
\label{tab:heldout_progress}
\centering\setlength{\tabcolsep}{2pt}
\AddToHookNext{env/tabular/begin}{\color{black}}
\begin{tabular}{lcc}
\toprule
\multirow{2}{*}{Data} & Initial-state SR & Unseen SR \\
 & (\%) & (\%) \\
\midrule
Mixed & 74 & 58 \\
CTR-hole & 92 & 74 \\
CTR-full & 94 & 76 \\
\bottomrule
\end{tabular}

\end{minipage}
\setcounter{table}{2}
\vspace{-4pt}
\end{table}

\paragraph{Behavior under task-required constraints.}
{
We directly test required cross-arm dependence with $\pi_{0.5}$ on two real-robot tasks. In Put Object Cabinet, we close the drawer just before the left arm inserts the object; attempting insertion while the drawer is closed is a violation, and the correct response is to wait until it is reopened. CTR waits in every trial, whereas the $L\!\rightarrow\!R$ baseline, which never observes waiting in front of the drawer, violates the constraint in 98\% of trials. In Organize Hardware, we define a shared workspace, where entry by one arm while the other occupies the region is a violation; CTR reduces violations from 36\% to 4\% against the Concurrent baseline (\paperref{Table}{tab:constraint_interventions}). Release latency increases from 0.01 to 0.27\,s and from 0.09 to 0.15\,s, respectively. CTR reduces constraint violations at the cost of longer release latency. Together, the Deliver Water probe, where CTR retains greater sensitivity during \emph{Serve Water} than in the independent phases, and these constraint interventions show phase-dependent sensitivity: reduced response when relative timing is irrelevant and retained response when coordination is required.
\looseness=-1\par
}

\begin{table}[t]
\caption{Task-constraint interventions with $\pi_{0.5}$ on two real-robot tasks ($n=50$). Put Object Cabinet compares $L\!\rightarrow\!R$ with CTR; Organize Hardware compares Concurrent with CTR. Lower is better for both violation rate and release latency; CTR lowers violations but has longer release latencies.}
\label{tab:constraint_interventions}
\centering
\AddToHookNext{env/tabular/begin}{\color{black}}
\begin{tabular}{llrrrr}
\toprule
\multirow{2}{*}[-0.4ex]{Task} & \multirow{2}{*}[-0.4ex]{Baseline} & \multicolumn{2}{c}{Violation (\%)} & \multicolumn{2}{c}{Release latency (s)} \\
\cmidrule(lr){3-4}\cmidrule(lr){5-6}
 & & Baseline & CTR & Baseline & CTR \\
\midrule
Put Object Cabinet (real) & $L\!\rightarrow\!R$ & 98 & 0 & 0.01 & 0.27 \\
Organize Hardware & Concurrent & 36 & 4 & 0.09 & 0.15 \\
\bottomrule
\end{tabular}

\vspace{-4pt}
\end{table}

\paragraph{Default-condition task performance.}
{
We next check whether the benefit extends to ordinary task execution across all eight tasks and three policy families. \paperref{Table}{tab:generality} compares CTR with the Serial baseline and shows higher task-averaged SR for each model family, in both simulation and real-robot evaluation. CTR improves success rates in 19 of 24 task--model comparisons, but does not improve performance on Place Dual Shoes.
\looseness=-1\par
}

\begin{table}[htbp]
\caption{Default-condition SRs (\%) for DP, ACT, and $\pi_{0.5}$ trained on Serial and CTR data. The first four rows are simulated tasks and the last four are real-robot tasks. Each entry is task-level SR ($n=50$ per task, model, and training condition).}
\label{tab:generality}
\centering\setlength{\tabcolsep}{4pt}
\AddToHookNext{env/tabular/begin}{\color{black}}
\begin{tabular}{lrrrrrr}
\toprule
\multirow{2}{*}[-0.4ex]{Task} & \multicolumn{2}{c}{DP} & \multicolumn{2}{c}{ACT} & \multicolumn{2}{c}{$\pi_{0.5}$} \\
\cmidrule(lr){2-3}\cmidrule(lr){4-5}\cmidrule(lr){6-7}
 & Serial & CTR & Serial & CTR & Serial & CTR \\
\midrule
Sort Blocks & 70 & 72 & 90 & 92 & 94 & 94 \\
Place Dual Shoes & 0 & 0 & 26 & 20 & 38 & 38 \\
Scan Object & 0 & 2 & 4 & 10 & 62 & 70 \\
Put Object Cabinet (sim) & 2 & 4 & 8 & 24 & 36 & 66 \\
\midrule
Sort Letters & 0 & 2 & 12 & 18 & 92 & 94 \\
Organize Hardware & 0 & 2 & 0 & 0 & 8 & 16 \\
Deliver Water & 20 & 26 & 14 & 34 & 78 & 92 \\
Put Object Cabinet (real) & 32 & 38 & 54 & 68 & 92 & 98 \\
\bottomrule
\end{tabular}

\vspace{-4pt}
\end{table}

\subsection{Generalization beyond training support}
\label{sec:exp-generalization}
To distinguish generalization from coverage of the tested bimanual progress states, we test an execution-time intervention absent from training and then explicitly withhold a region of the bimanual progress-state space.

\paragraph{Unseen execution delays.}
{
On the Independent Sort Letters task, we externally pause only the opposite arm for $d\in\{1,2,3,4\}$ seconds, matching the evaluated arm's local initial state between control and intervention branches. Since this pause does not change the evaluated arm's task requirements, selective dependence predicts little transfer to its completion time.
\looseness=-1\par
}

\begingroup
\Needspace{0.42\textheight}
\setlength{\intextsep}{3pt}
\begin{wrapfigure}[23]{r}{0.45\textwidth}
\centering
\includegraphics[width=\linewidth]{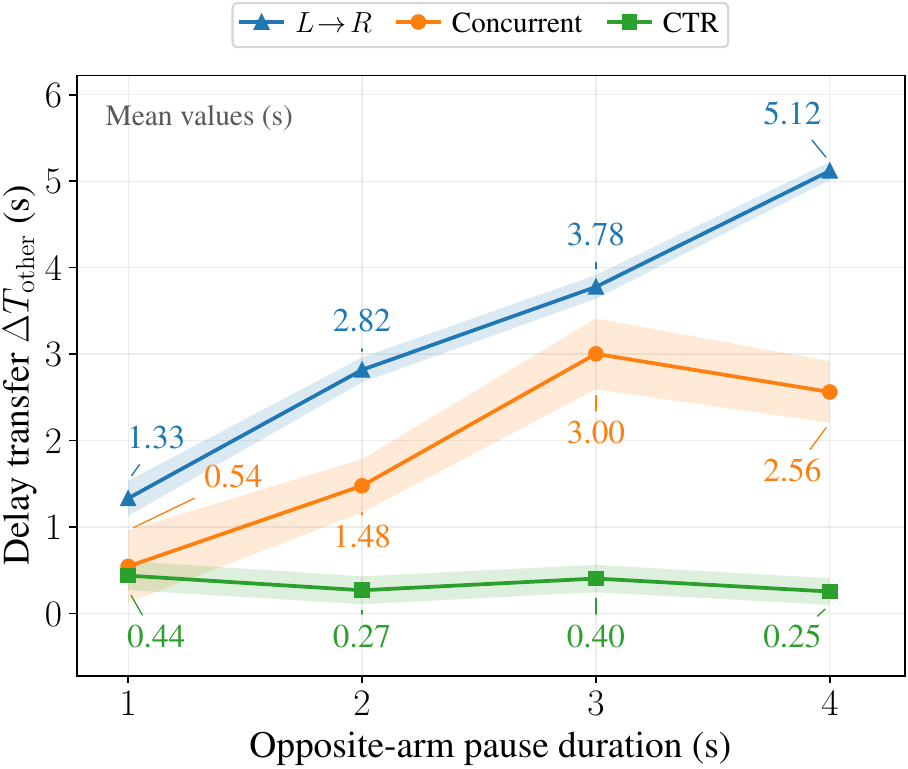}
\caption{Opposite-arm pause on Independent Sort Letters with $\pi_{0.5}$. Delay transfer is the unpaused arm's extra completion time relative to a matched no-pause control. Numbers show means (s); faint bands show pointwise 95\% confidence intervals (50 trials per condition and duration). Straight segments join only the four tested durations.}
\label{fig:pause-delay-ci}
\end{wrapfigure}
\noindent We measure
$\Delta T_{\rm other}(d)=T_{\rm other}^{\rm pause}(d)-T_{\rm other}^{\rm control}$
and fit its slope $\beta_{\rm delay}$ across pause durations; slopes near one and zero indicate near-complete transfer and approximate invariance, respectively. For $\pi_{0.5}$, a 4\,s pause transfers 5.12\,s under $L\!\rightarrow\!R$, 2.56\,s under Concurrent, and 0.25\,s under CTR, with fitted slopes of 1.232, 0.758, and $-0.042$ (\paperref{Figure}{fig:pause-delay-ci}). CTR's weak response to irrelevant opposite-arm progress thus persists under an external execution delay.
\looseness=-1\par

\paragraph{Held-out progress-state region.}
Because an unseen pause can still visit progress states covered by CTR, we also withhold progress states from training. On Sort Letters, we normalize each arm's progress within a reschedulable block to $[0,1]$ and fix one held-out region, $H=[0.20,0.30]\times[0.70,0.80]$, leaving both arm-local progress values individually familiar. Mixed retains the three-schedule mixture, CTR-hole removes training samples whose bimanual progress states lie in $H$, and CTR-full retains full admissible support; training budgets are otherwise matched. We evaluate $\pi_{0.5}$ from both the standard initial state and states inside $H$. CTR-hole achieves 74\% unseen-state SR versus 58\% for Mixed and 76\% for CTR-full; their standard initial-state SRs are 92\%, 74\%, and 94\%, respectively (\paperref{Table}{tab:heldout_progress}). The improvement over Mixed despite withholding $H$ supports generalization to unobserved bimanual progress states.
\looseness=-1\par
\endgroup

\setcounter{table}{5}

\section{Limitations}
{
\method{} relies on the task-specific processing pipelines; errors in interaction detection, segment boundaries, or task-relation logic can create invalid recombinations. Main-view composition only approximates a physically realized concurrent scene, so contact-sensitive and coupled segments remain protected rather than independently recombined. Finally, our interventions and diagnostics cover specific forms of cross-arm dependence rather than every possible bimanual interaction.
\looseness=-1\par
}

\section{Conclusion}
{
A bimanual policy should learn the task, not the demonstrator's schedule: it should remain invariant to the relative timing of independently schedulable arm-local subtasks while preserving dependence imposed by task-required cross-arm constraints. \method{} promotes this selective cross-arm dependence by exposing the same arm-local sensorimotor content to diverse admissible schedules. Across our diagnostics and interventions, broader admissible schedule support suppresses spurious cross-arm coupling while retaining task-required coordination.
\looseness=-1\par
}
\label{sec:mainend}

\section*{Reproducibility statement}
We submit an anonymized code package and the datasets generated with \method{} as supplementary material with this paper. The code release provides the implementation needed to construct the \method{} training data and reproduce the experimental workflow, while the accompanying generated datasets make the outputs of our data-construction procedure directly available for inspection and use without requiring reviewers to rerun the full generation pipeline. \paperref{Appendix}{app:training-evaluation-config} reports the matched training and simulation-evaluation configurations, \paperref{Appendix}{app:metrics} defines the reported metrics and aggregation rules, and \paperref{Appendix}{app:implementation} specifies the \method{} data-construction procedure, including segmentation and constraint extraction, admissible-schedule construction, temporal recombination, IdleMask supervision, qualification checks, and provenance/rejection handling. The supplementary code package is anonymized for double-blind review.

\section*{Ethics statement}
Temporal recombination is applied only within task-annotated admissible regions; prerequisites, sustained support, shared-object interactions, and workspace exclusions remain protected. Real-robot deployment uses the same independent safety procedures as the baseline policies. Any release of recorded camera streams should respect consent, privacy, and applicable data-use permissions.

\section*{AI use statement}
{
GPT-6 Astra was used at task setup time to generate task-specific video-processing pipelines. Those programs---not GPT-6 Astra itself---process the recorded demonstrations to detect task-relevant interactions, segment arm-local trajectories, produce the temporal dependency metadata that separates reschedulable from protected relations, and execute the task-specific recombination and validation logic. GPT-6 Astra is therefore not used as a per-video annotator and is not queried for episode-level labels. Generative AI was also used to discuss hypotheses and experimental design, assist literature discovery and comparison, formalize the method, analyze aggregate results, and draft and edit the manuscript and LaTeX. The robot experiments and reported empirical outcomes are produced by the experimental pipeline rather than by the writing model.
\looseness=-1\par
}

\bibliographystyle{iclr2027_conference}
\bibliography{references}
\clearpage
\appendix
\section{Survey of schedule coverage in existing bimanual datasets}
\label{app:dataset_survey}

{
We survey \surveyDatasets{} publicly available bimanual manipulation datasets:
RoboTwin~2.0~\citep{chen2025robotwin2},
RoboCOIN~\citep{wu2025robocoin},
DexMimicGen~\citep{jiang2025dexmimicgen}, and
BiCoord~\citep{peng2026bicoord}
(\paperref{Table}{tab:survey-datasets}). A task is \emph{reschedulable} when the relative timing
or ordering of its two arms can change without altering task semantics or violating a
task-required dependency. We call a reschedulable task \emph{diversity-covered} when its
released demonstrations contain at least two materially different schedules; otherwise
it is \emph{locked}. Different schedules differ in subtask order or in a material
inter-arm timing offset.
\looseness=-1\par
}

\begin{table}[ht]
\caption{Dataset size and schedule coverage for the surveyed bimanual datasets. Locked is the percentage of reschedulable tasks demonstrated under a single observed schedule.}
\label{tab:survey-datasets}
\centering\setlength{\tabcolsep}{4pt}
\AddToHookNext{env/tabular/begin}{\color{black}}
\begin{tabular}{llrrrr}
\toprule
\multirow{2}{*}{Dataset} & \multirow{2}{*}{Domain} & Released & Reschedulable & Diversity-covered & Locked \\
        &        & tasks & tasks & tasks & (\%) \\
\midrule
RoboTwin 2.0 & Simulation & 50 & 12 & 0 & 100.0 \\
RoboCOIN & Real & 639 & 231 & 0 & 100.0 \\
DexMimicGen & Simulation & 9 & 8 & 6 & 25.0 \\
BiCoord & Simulation & 18 & 15 & 3 & 80.0 \\
\midrule
\multicolumn{2}{l}{Pooled} & 716 & 266 & 9 & 96.6 \\
\bottomrule
\end{tabular}

\end{table}

{
Across the \surveyDatasets{} datasets, the survey covers \surveyTasks{} released tasks, of
which \surveyReschedulable{} (\surveyReschedulablePct{}\%) are reschedulable. Only
\surveyDiverse{} of those tasks (\surveyDiversePct{}\%) are diversity-covered, leaving
\surveyLockedPct{}\% locked to a single observed schedule. The pattern is present in
every dataset: RoboTwin~2.0 and RoboCOIN are 100\% locked, BiCoord is 80\% locked, and
DexMimicGen is 25\% locked. The equal-weighted mean of the four per-dataset locked
fractions is \surveyMacroLockedPct{}\%, showing that the conclusion is not driven only by
the larger RoboCOIN collection. Overall, observed schedule coverage is narrow across
existing bimanual datasets.
\looseness=-1\par
}

\section{Task and evaluation inventory}
\label{app:tasks}
{
The evaluation covers four simulated and four real-robot tasks spanning Independent, Shared-Workspace Exclusion, Coordination, and Prerequisite settings. Independent relations permit either arm's local progress to vary without affecting the other; Shared-Workspace Exclusion requires only non-overlap in a conflict region; Coordination requires a coupled or synchronized bimanual phase; and Prerequisite requires one arm to establish a condition before the other may proceed.
\looseness=-1\par
}

\begin{table}[ht]
\caption{Task inventory used in the experiments.}
\label{tab:tasks}
\centering
\AddToHookNext{env/tabularx/begin}{\color{black}}
\begin{tabularx}{\linewidth}{llXl}
\toprule
ID & Domain & Task & Class \\
\midrule
S1 & Sim. & Sort Blocks & Independent \\
S2 & Sim. & Place Dual Shoes & Shared-Workspace Exclusion \\
S3 & Sim. & Scan Object & Coordination \\
S4 & Sim. & Put Object Cabinet & Prerequisite \\
R1 & Real & Sort Letters & Independent \\
R2 & Real & Organize Hardware & Shared-Workspace Exclusion \\
R3 & Real & Deliver Water & Coordination \\
R4 & Real & Put Object Cabinet & Prerequisite \\
\bottomrule
\end{tabularx}
\end{table}

\paragraph{Why each task has this relation.}
{
\textbf{S1, Sort Blocks (Independent):} this task is adapted from the Blocks Ranking task in RoboTwin~2.0~\citep{chen2025robotwin2}. We modify the setup so that each side contains two blocks to be manipulated, giving the two arms the same number of arm-local subtasks. The arms operate on disjoint block sets, and either arm can start, progress, or finish its local sorting without changing the validity of the other arm's actions. \textbf{R1, Sort Letters (Independent):} the two arms sort distinct letters into arm-local destinations, so one arm's progress is neither a precondition nor a synchronization signal for the other.
\looseness=-1\par
}

{
\textbf{S2, Place Dual Shoes (Shared-Workspace Exclusion):} the two placement motions are locally executable by either arm, but they use an overlapping placement/workspace region; simultaneous occupancy is therefore disallowed even though the motions outside that conflict region can be rescheduled. \textbf{R2, Organize Hardware (Shared-Workspace Exclusion):} the arms manipulate separate hardware items but share a conflict region, so one arm must wait while the other occupies that region and may proceed once it is released.
\looseness=-1\par
}

{
\textbf{S3, Scan Object (Coordination):} successful scanning contains a coupled phase in which the two arms must jointly establish and maintain the required scan configuration, so that phase cannot be independently retimed across arms. \textbf{R3, Deliver Water (Coordination):} the \emph{Serve Water} phase requires mutually timed bimanual actions, while the surrounding \emph{Place Bottles} and \emph{Return to Standby} phases remain independently schedulable.
\looseness=-1\par
}

{
\textbf{S4, Put Object Cabinet (Prerequisite):} one arm must establish cabinet access before the other arm can validly insert or place the object, imposing a required order rather than a mere scheduling preference. \textbf{R4, Put Object Cabinet (Prerequisite):} the dependent placement must likewise wait for the cabinet-ready condition; withholding that condition should block the dependent arm until the prerequisite is released.
\looseness=-1\par
}

\paragraph{Progress-state interventions.}
{
For the progress-state interventions, each arm's local progress is normalized to $[0,1]$. On Sort Letters, the six evaluated states are C00 $(0,0)$, Cmid $(0.5,0.5)$, L-lead $(0.75,0.25)$, R-lead $(0.25,0.75)$, CL $(1,0)$, and CR $(0,1)$. On Put Object Cabinet, the matched suite is C00 $(0,0)$, Cmid $(0.5,0.5)$, L-lead $(0.75,0.25)$, R-lead $(0.25,0.5)$, CL $(1,0)$, and CR $(0,0.5)$. C00 is the nominal start, Cmid is matched progress, L-lead and R-lead are interior off-canonical progress states, and CL/CR are task-specific ready or completion-side states.
\looseness=-1\par
}

\paperref{Figure}{fig:progress-initial-states} shows the frozen initial scenes used for these two six-state evaluations.

\begin{figure}[htbp]
\centering
\includegraphics[width=\linewidth]{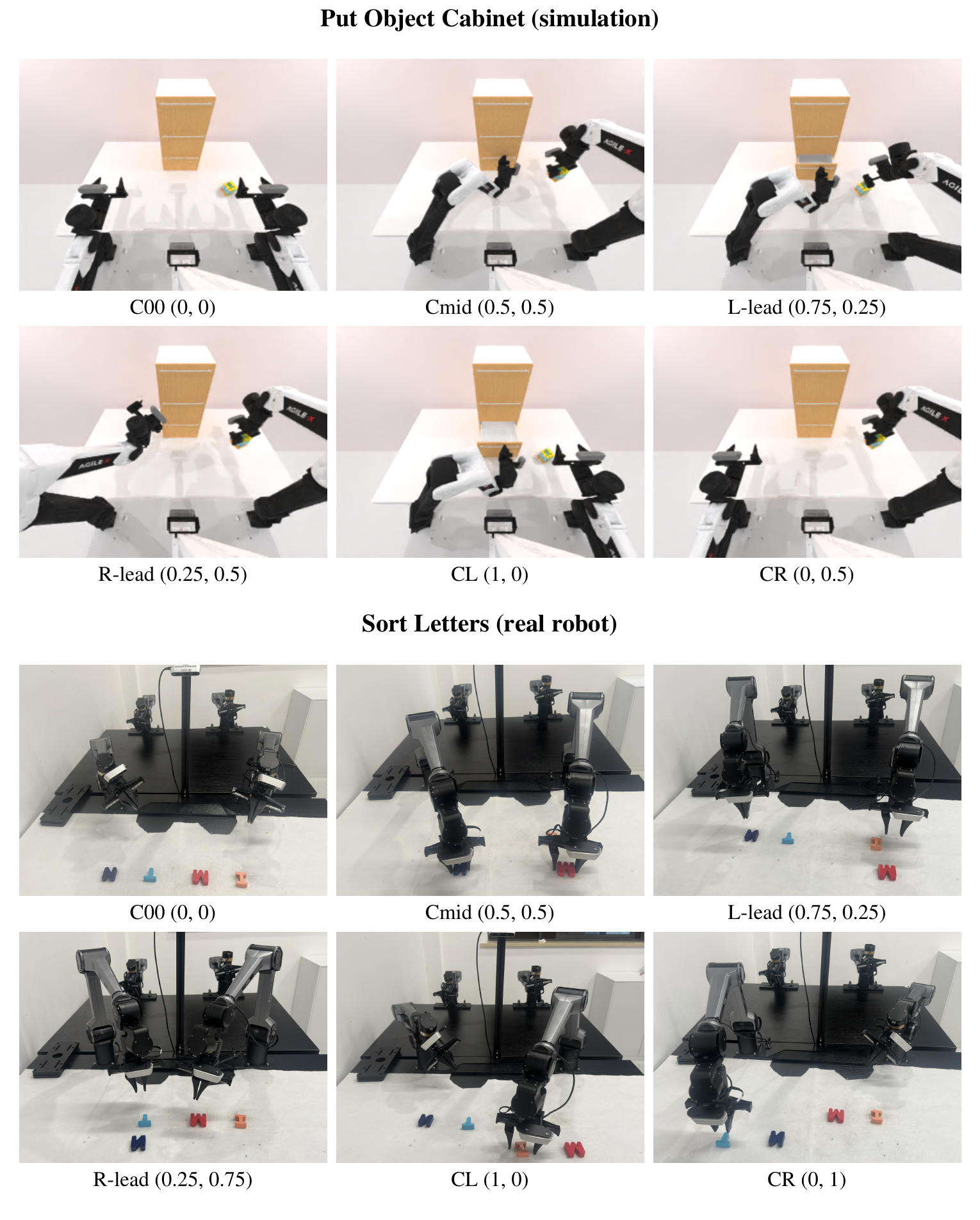}
\caption{Initial-scene setups for the six progress-state evaluations. The first two rows show Put Object Cabinet in simulation; the last two rows show Sort Letters on the real robot. Each panel is labeled by its setup name and normalized bimanual progress state.}
\label{fig:progress-initial-states}
\end{figure}

\section{Training and simulation-evaluation configurations}
\label{app:training-evaluation-config}

\paragraph{Demonstrations and frames.}
In the 50-demonstration setting, each task has 50 independent source episodes. Temporal recombination can produce multiple derived episodes from one source. Their durations and recorded frame counts vary with the task and schedule; training uses the valid, unpadded frames rather than forcing equal frame counts.

\paragraph{Unified training protocol.}
For the checkpoints reported in this paper, all $\pi_{0.5}$ models use the OpenPI implementation~\citep{openpi2025} with LoRA~\citep{hu2021lora}: simulation models use 10,000 optimizer steps and real-robot models use 20,000 steps. All ACT models in both domains use batch size 8 and 100,000 optimizer steps. All DP models in both domains use 600 effective epochs, with each epoch defined over eligible training frame-window anchors.

\paragraph{Simulation evaluation.}
{
Within each task, all compared policies use the same frozen initial-scene suite, camera calibration, control clock, native action limit, success predicate, and rollout timeout. The common rollout timeout is used only to terminate unsuccessful evaluations; it is not the failure penalty used in $T_{\rm Pen}$. Model-specific action queues and inference procedures are retained; parallel workers change only execution speed.
\looseness=-1\par
}

\section{Controlled schedule distributions: schematic illustrations}
\label{app:schedule-distributions}
{
\paperref{Figure}{fig:schedule-schematics} illustrates the geometry of the five controlled schedule distributions used in the main text. $L\!\rightarrow\!R$ contains only left-first serial execution; Serial combines $L\!\rightarrow\!R$ and $R\!\rightarrow\!L$; Concurrent starts both arms together; Mixed combines both serial orders and Concurrent; and \method{} spans intermediate admissible schedules. For the real-robot tasks, the $L\!\rightarrow\!R$ distribution corresponds to the original, unretimed demonstrations; in simulation, we construct the same $L\!\rightarrow\!R$ schedule as the matched reference distribution. The curves are schematic, do not represent a particular physical task, and do not generate the empirical observations in \paperref{Figure}{fig:schedule-distributions}. Each schematic trajectory has independently normalized time; the number of displayed curves does not encode the distribution's mixture weights.
\looseness=-1\par
}

\begin{figure}[htbp]
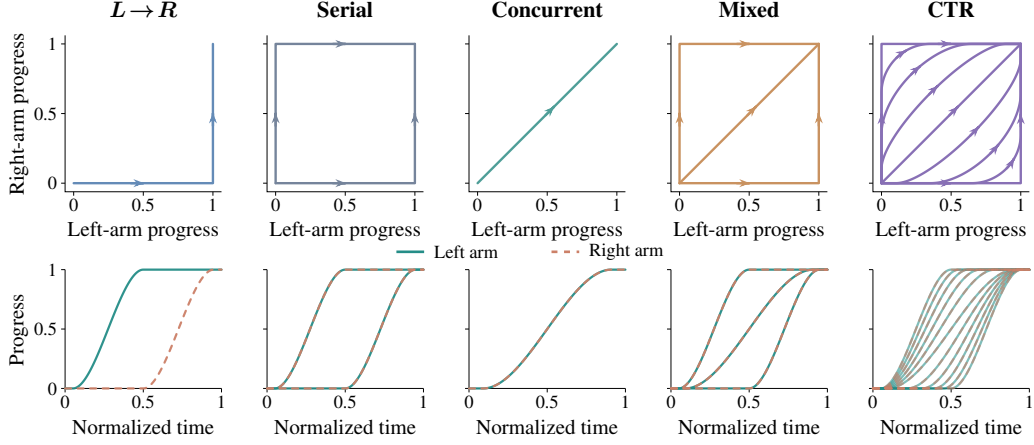

\centering
\begingroup
\setlength{\unitlength}{1bp}
\resizebox{\linewidth}{!}{%
\begin{picture}(1600,680)
\put(0,0){
\begin{tikzpicture}[x=1bp,y=1bp]
\path[use as bounding box] (0,0) rectangle (1600,305);
\definecolor{leftarm}{HTML}{248D87}
\definecolor{rightarm}{HTML}{CC8068}
\draw[leftarm,line width=4bp] (590,287) -- (632,287);
\node[anchor=west,inner sep=0,font=\fontsize{28}{34}\selectfont] at (647,287) {Left arm};
\draw[rightarm,line width=4bp,dash pattern=on 9bp off 10bp] (825,287) -- (867,287);
\node[anchor=west,inner sep=0,font=\fontsize{28}{34}\selectfont] at (882,287) {Right arm};
\foreach \i in {0,...,4} {
  \begin{scope}[shift={(88+306*\i,80)}]
    \draw[line width=1.5bp] (0,180) -- (0,0) -- (237,0);
    \foreach \v in {0,0.5,1} {
      \draw[line width=1.5bp] ({237*\v},0) -- ++(0,-6);
      \node[anchor=north,inner sep=0,font=\fontsize{28}{34}\selectfont] at ({237*\v},-10) {\v};
      \draw[line width=1.5bp] (0,{180*\v}) -- ++(-6,0);
      \ifnum\i=0
        \node[anchor=east,inner sep=0,font=\fontsize{28}{34}\selectfont] at (-12,{180*\v}) {\v};
      \fi
    }
    \node[inner sep=0,font=\fontsize{32}{38}\selectfont] at (118.5,-59) {Normalized time};
    \begin{scope}[x=237bp,y=180bp,every path/.style={line width=3.4bp,line cap=round,line join=round}]
      \input{figures/schedule_comparison/arm-progress-paths.tex}
    \end{scope}
  \end{scope}
}
\end{tikzpicture}%
}
\put(0,345){
\begin{tikzpicture}[x=1bp,y=1bp]
\path[use as bounding box] (0,0) rectangle (1600,285);
\foreach \i/\shade in {0/648BB7,1/75849B,2/4F9E98,3/C99260,4/8870B5} {
  \definecolor{schedulecurve}{HTML}{\shade}
  \begin{scope}[shift={(88+306*\i,33)}]
    \draw[line width=1.5bp] (0,237) -- (0,0) -- (237,0);
    \foreach \v in {0,0.5,1} {
      \draw[line width=1.5bp] ({13+211*\v},0) -- ++(0,-6);
      \node[anchor=north,inner sep=0,font=\fontsize{28}{34}\selectfont]
        at ({13+211*\v},-10) {\v};
      \draw[line width=1.5bp] (0,{13+211*\v}) -- ++(-6,0);
      \ifnum\i=0
        \node[anchor=east,inner sep=0,font=\fontsize{28}{34}\selectfont]
          at (-12,{13+211*\v}) {\v};
      \fi
    }
    \begin{scope}[shift={(13,13)},x=211bp,y=211bp,
      every path/.style={draw=schedulecurve,line width=3.6bp,line cap=round,line join=round}]
      \input{figures/schedule_comparison/joint-progress-paths.tex}
    \end{scope}
  \end{scope}
}
\end{tikzpicture}%
}
\put(206.5,655){\makebox(0,0){\fontsize{34}{40}\selectfont\bfseries\boldmath $L\!\rightarrow\!R$}}
\put(512.5,655){\makebox(0,0){\fontsize{34}{40}\selectfont\bfseries\boldmath Serial}}
\put(818.5,655){\makebox(0,0){\fontsize{34}{40}\selectfont\bfseries\boldmath Concurrent}}
\put(1124.5,655){\makebox(0,0){\fontsize{34}{40}\selectfont\bfseries\boldmath Mixed}}
\put(1430.5,655){\makebox(0,0){\fontsize{34}{40}\selectfont\bfseries\boldmath \method{}}}
\put(18,496){\rotatebox{90}{\makebox(0,0){\fontsize{32}{38}\selectfont Right-arm progress}}}
\put(18,169){\rotatebox{90}{\makebox(0,0){\fontsize{32}{38}\selectfont Progress}}}
\put(206.5,319){\makebox(0,0){\fontsize{32}{38}\selectfont Left-arm progress}}
\put(512.5,319){\makebox(0,0){\fontsize{32}{38}\selectfont Left-arm progress}}
\put(818.5,319){\makebox(0,0){\fontsize{32}{38}\selectfont Left-arm progress}}
\put(1124.5,319){\makebox(0,0){\fontsize{32}{38}\selectfont Left-arm progress}}
\put(1430.5,319){\makebox(0,0){\fontsize{32}{38}\selectfont Left-arm progress}}
\end{picture}%
}
\endgroup
\caption{Idealized schedule geometry. Top: idealized bimanual progress-state support with left-arm progress on the horizontal axis and right-arm progress on the vertical axis. Arrows indicate increasing progress. Axis padding separates the trajectories from the axes without changing their shapes; in the top-row \method{} panel, the two interior paths nearest the serial envelope are omitted for clarity. Bottom: the corresponding schedule families as arm progress over independently normalized time; solid teal and dashed orange denote the left and right arms, respectively. Both rows are schematic.}
\label{fig:schedule-schematics}
\end{figure}

\paragraph{Empirical onset measurement.}
{
The densities in \paperref{Figure}{fig:schedule-distributions} use Sort Blocks simulation trajectories. An arm's onset is the first of three consecutive recorded frames whose end-effector translation exceeds 2\,mm from its initial position. Thus $\Delta_{\mathrm{start}}=t_R^{\rm start}-t_L^{\rm start}>0$ means that the left arm starts first; a negative value means that the right arm starts first. $L\!\rightarrow\!R$ contains 50 left-first episodes; each other group contains 100 episodes. Serial contains 50 episodes per order, and Mixed contains 34 left-first, 33 right-first, and 33 Concurrent episodes. Groups share source scenes and some episodes. Histograms use shared 1\,s bins, with density equal to count divided by sample size and bin width, so each histogram has unit area. Gaussian kernel density estimates use a fixed 0.6\,s bandwidth; their tails reflect smoothing rather than additional observed support. Concurrent remains unsmoothed because all observed offsets are zero at the approximately 0.04\,s recording resolution. Vertical scales differ across panels.
\looseness=-1\par
}

\section{Metric definitions and aggregation}
\label{app:metrics}

\subsection{Success and failure-penalized completion time}
For each policy--state cell $c$, let $N_c$ be the number of evaluations, $S_{cj}\in\{0,1\}$ indicate success, and $t_{cj}^{\rm goal}$ be the completion time for successful evaluation $j$. When $n_c^+=\sum_j S_{cj}>0$, define the cell's mean successful completion time and failure penalty as
\begin{equation}
 \bar t_c^+=\frac{1}{n_c^+}\sum_{j=1}^{N_c}S_{cj}t_{cj}^{\rm goal},
 \qquad P_c=2\bar t_c^+.
\end{equation}
We then report
\begin{equation}
 \mathrm{SR}_c=100\frac{n_c^+}{N_c},\qquad
 T_{{\rm Pen},c}=\frac{1}{N_c}\sum_{j=1}^{N_c}\left[S_{cj}t_{cj}^{\rm goal}+(1-S_{cj})P_c\right].
\end{equation}
Thus each failed evaluation receives a penalty equal to twice the mean successful completion time of that same cell, rather than the common rollout timeout. Equivalently, with success proportion $s_c=n_c^+/N_c$, $T_{{\rm Pen},c}=(2-s_c)\bar t_c^+$. If $n_c^+=0$, $\bar t_c^+$ and $T_{{\rm Pen},c}$ are undefined and we report ``--''.

\paragraph{Undefined completion-time entries.}
For cells with SR=0, \paperref{Table}{tab:progress_robustness} reports ``--'' because the cell has no successful completion times from which to define $\bar t_c^+$ and hence $T_{{\rm Pen},c}$.

\paragraph{Aggregation convention.}
Unless otherwise stated, summaries across tasks, model families, or conditions are equal-weighted macro-averages of the reported condition-level values, computed before display rounding. In particular, the default-condition ``overall,'' simulation/real, and model-family summaries use equal-weighted means of task-level SRs rather than pooling episodes across heterogeneous tasks.

\subsection{Relative start time}
Schedule inheritance is measured by
\begin{equation}
 \Delta_{\rm start}=t_R^{\rm start}-t_L^{\rm start}.
\end{equation}
Condition-level entries report the arithmetic mean of $\Delta_{\rm start}$ over evaluations in that condition.

\begin{table}[htbp]
\caption{Schedule inheritance on Sort Letters. Policy entries are mean $\Delta_{\rm start}=t_R^{\rm start}-t_L^{\rm start}$ in seconds. The Demos average row reports dataset-average demonstration references: $+17.10$ s for $L\!\rightarrow\!R$ (mean left-arm task duration), $0.00$ s for Concurrent, and $-17.22$ s for $R\!\rightarrow\!L$ (negative mean right-arm task duration).}
\label{tab:schedule_inheritance}
\centering
\AddToHookNext{env/tabular/begin}{\color{black}}
\begin{tabular}{lccc}
\toprule
Source & $L\!\rightarrow\!R$ & Concurrent & $R\!\rightarrow\!L$ \\
\midrule
\textbf{Demos average} & 17.10 & 0.00 & $-17.22$ \\
\midrule
DP & 14.74 & 0.05 & $-15.15$ \\
ACT & 20.31 & $-0.04$ & $-19.78$ \\
$\pi_{0.5}$ & 14.52 & 0.02 & $-14.43$ \\
\bottomrule
\end{tabular}
\end{table}

\subsection{Phase-conditioned cross-arm sensitivity}
\label{app:phase-sensitivity-probe}
\begin{figure*}[t]
\centering
\includegraphics[width=\linewidth]{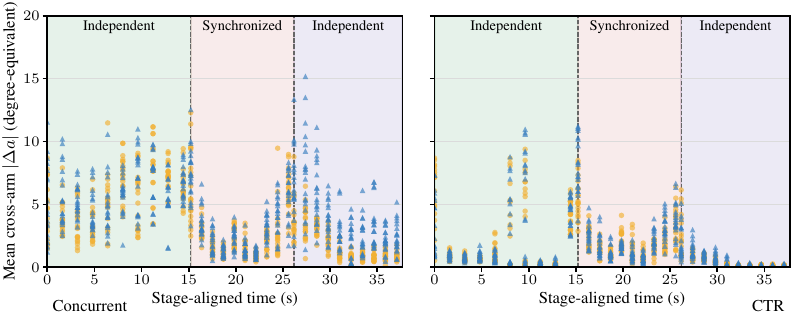}
\caption{Phase-conditioned cross-arm sensitivity for $\pi_{0.5}$ 20k checkpoints trained on Concurrent data (left) and CTR data (right). Yellow circles measure right-state-to-left-Action sensitivity; blue triangles measure left-state-to-right-Action sensitivity. The two outer regions are labeled Independent and the middle region is Synchronized.}
\label{fig:phase-sensitivity-pi05}
\end{figure*}

{
We use the 20-episode StagePair dataset for Deliver Water, selecting 600 matched groups: 300 with the left arm fixed and 300 with the right arm fixed. The groups are balanced across \emph{Place Bottles}, \emph{Serve Water}, and \emph{Return to Standby}, with 100 groups per phase and intervention direction. Each group contributes two counterfactual points. In the $L\!\leftarrow\!R$ direction, the right-arm state is changed while the left-arm local state is preserved, and we measure the resulting change in the predicted left-arm Action. In the $R\!\leftarrow\!L$ direction, the left-arm state is changed while the right-arm local state is preserved, and we measure the resulting change in the predicted right-arm Action. Each phase therefore has 200 plotted points per direction. We compare matched CTR and Concurrent checkpoints for ACT, DP, and $\pi_{0.5}$; \paperref{Figure}{fig:phase-conditioned-sensitivity} summarizes all three with mean curves, and \paperref{Figures}{fig:phase-sensitivity-act}--\ref{fig:phase-sensitivity-pi05} show the directional scatter plots here. All composite scores use the shared gripper factor $0.7608885$ degree-equivalent/mm described below.
\looseness=-1\par
}

{
For evaluated arm $i$ and intervened opposite arm $j$, let $a_{k,h,d}^{i,\mathrm{ref}}$ and $a_{k,h,d}^{i,\mathrm{cf}(j)}$ be the reference and counterfactual predictions for matched group $k$, prediction step $h$, and joint Action dimension $d$.
The six joint Action coordinates are measured in degrees and the gripper coordinate in millimeters. To avoid averaging these unlike units directly, we map gripper changes to a shared degree-equivalent scale using robust spans from the StagePair state distribution. Let $Q_{p,d}$ be percentile $p$ of state coordinate $d$. Across the twelve joint coordinates in both arms, the median $Q_{90,d}-Q_{10,d}$ is $s_J=42.9902^\circ$; across the two gripper coordinates, the median span is $s_g=56.5000$ mm. We use the same factor for every checkpoint and direction:
\looseness=-1\par
}
\begin{equation}
 c_g=\frac{s_J}{s_g}=0.7608885\;\mathrm{degree\text{-}equivalent/mm},
 \qquad
 |\Delta g|_{\rm eq}=c_g|\Delta g_{\rm mm}|.
 \label{eq:gripper-normalization}
\end{equation}
This is a shared scale convention based on the input-state ranges, not a claim that gripper travel is physically an angle. The factor and source quantiles are recorded in the probe's normalization metadata.

Using six joint dimensions and one normalized gripper dimension, the point-level directional sensitivity is
\begin{equation}
 d_{k,m}^{i\leftarrow j}
 =
 \frac{1}{7H_m}
 \sum_{h=1}^{H_m}\left[
 \sum_{d=1}^{6}\left|
 a_{k,h,d}^{i,\mathrm{cf}(j)}-a_{k,h,d}^{i,\mathrm{ref}}
 \right|
 +c_g\left|
 g_{k,h}^{i,\mathrm{cf}(j)}-g_{k,h}^{i,\mathrm{ref}}
 \right|\right],
 \qquad H_m\in\{8,50\}.
 \label{eq:directional-sensitivity-point}
\end{equation}
{
The resulting metric is in degree-equivalent units. ACT and $\pi_{0.5}$ use 50-step predicted chunks; DP uses the eight executed steps from its 16-step sampling horizon. Within each checkpoint pair, both scatter panels use a shared y-axis. For visualization, source episodes are aligned within each phase by linearly scaling each phase duration to the mean duration of that phase; this changes only the horizontal coordinate of a point, not its sensitivity value. The main-text curves divide each aligned phase into ten equal-width time bins. Each bin contains 20 points per intervention direction; we average within each direction and then average the two directional means, placing the result at the mean aligned time of the bin's points. No additional smoothing is applied, and line segments do not cross phase boundaries. All three main-text panels share one y-axis scale. In the plots, \emph{Place Bottles} and \emph{Return to Standby} are labeled \emph{Independent}; \emph{Serve Water} is labeled \emph{Synchronized}.
\looseness=-1\par
}

Let $\mathcal K_p^{L\leftarrow R}$ and $\mathcal K_p^{R\leftarrow L}$ denote the points in phase $p\in\{\mathrm{place},\mathrm{serve},\mathrm{return}\}$ for the two intervention directions. We first average within each direction and then average the two directions:
\begin{equation}
 S_p
 =
 \frac{1}{2}
 \left[
 \frac{1}{|\mathcal K_p^{L\leftarrow R}|}
 \sum_{k\in\mathcal K_p^{L\leftarrow R}} d_{k,m}^{L\leftarrow R}
 +
 \frac{1}{|\mathcal K_p^{R\leftarrow L}|}
 \sum_{k\in\mathcal K_p^{R\leftarrow L}} d_{k,m}^{R\leftarrow L}
 \right].
 \label{eq:phase-sensitivity}
\end{equation}
Thus $S_{\rm place}$, $S_{\rm serve}$, and $S_{\rm return}$ are simply the phase-wise means of the two directional point clouds, with the two directions weighted equally even if their sample counts differ. The selectivity gap is
\begin{equation}
 G_{\rm sel}=S_{\rm serve}-\frac{1}{2}(S_{\rm place}+S_{\rm return}).
 \label{eq:appendix-selectivity-gap}
\end{equation}
When the two directions contain the same number of points in a phase, \paperref{Equation}{eq:phase-sensitivity} is equivalent to taking the mean y-value of all yellow and blue points in that phase. The probe is diagnostic rather than a task-success test.

\paperref{Table}{tab:phase-sensitivity-gsel} checks the selectivity gap after shortening the prediction horizon and after excluding chunks that reach beyond their annotated phase. The one-step and first-eight-step scores use the first 1 and 8 predicted Actions, respectively; the phase-contained score uses the full model chunk only when every predicted step remains within the sampled phase.

{
At 30~Hz, a 50-step ACT or $\pi_{0.5}$ chunk spans about 1.67~s. Of the 100 groups per direction and phase, 10 \emph{Place Bottles} groups per direction reach into \emph{Serve Water}, and 12 \emph{Serve Water} groups per direction reach into \emph{Return to Standby}. At the end of \emph{Return to Standby}, 17 groups in $L\!\leftarrow\!R$ and 14 in $R\!\leftarrow\!L$ extend beyond the labeled segment end; the strict phase-contained column excludes these as well. DP's eight-step chunk spans 0.27~s, and none of its sampled chunks crosses a labeled segment boundary.
\looseness=-1\par
}

\begin{table*}[t]
\caption{Normalized phase-selectivity gap $G_{\rm sel}$ in degree-equivalent units. Gripper differences are scaled by $0.7609$ degree-equivalent/mm using median $q_{90}-q_{10}$ spans of $42.9902^\circ$ for joints and $56.5000$ mm for grippers. A positive value means sensitivity is higher during \emph{Serve Water} than the mean of \emph{Place Bottles} and \emph{Return to Standby}. The phase-contained column excludes chunks that extend beyond the sampled phase boundary; the first-step and first-eight-step columns include all sampled groups.}
\label{tab:phase-sensitivity-gsel}
\centering\setlength{\tabcolsep}{5pt}
\begin{tabular}{llrrrr}
\toprule
Model & Training data & Full chunk & First step & First 8 steps & Phase-contained full chunk \\
\midrule
ACT & CTR & $+2.26$ & $+3.32$ & $+3.16$ & $+2.22$ \\
ACT & Concurrent & $-5.71$ & $-5.12$ & $-5.62$ & $-6.35$ \\
DP & CTR & $+1.60$ & $+1.14$ & $+1.60$ & $+1.60$ \\
DP & Concurrent & $-3.22$ & $-3.06$ & $-3.22$ & $-3.22$ \\
$\pi_{0.5}$ 20k & CTR & $+1.11$ & $+0.33$ & $+0.61$ & $+1.05$ \\
$\pi_{0.5}$ 20k & Concurrent & $-1.22$ & $+0.11$ & $0.00$ & $-1.63$ \\
\bottomrule
\end{tabular}
\end{table*}

{
The ACT and DP gaps retain their signs at one step, over eight steps, and on phase-contained full chunks. For $\pi_{0.5}$, the CTR gap remains positive across these checks, while Concurrent is near zero at the first step and over eight steps but negative for the full phase-contained chunk. This suggests the $\pi_{0.5}$ difference develops later in the predicted horizon. Removing boundary-reaching chunks changes the full-chunk gaps only modestly, so phase crossing alone does not explain the full-horizon pattern.
\looseness=-1\par
}

\begin{figure*}[t]
\centering
\includegraphics[width=\linewidth]{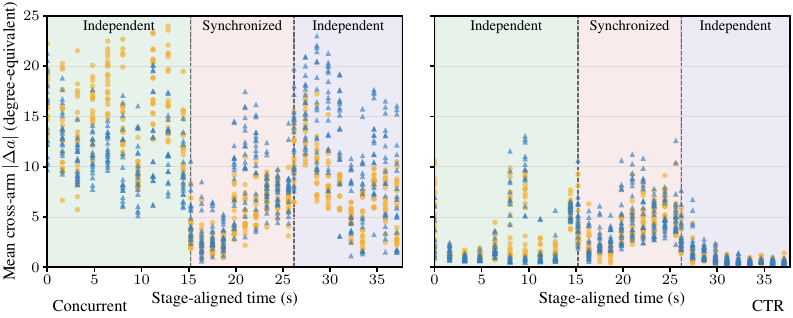}
\caption{Phase-conditioned cross-arm sensitivity for ACT checkpoints trained on Concurrent data (left) and CTR data (right). Yellow circles measure right-state-to-left-Action sensitivity; blue triangles measure left-state-to-right-Action sensitivity. The two outer regions are labeled Independent and the middle region is Synchronized.}
\label{fig:phase-sensitivity-act}
\end{figure*}

\begin{figure*}[t]
\centering
\includegraphics[width=\linewidth]{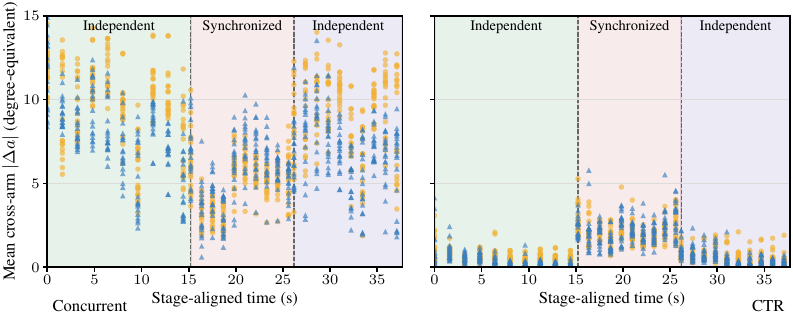}
\caption{Phase-conditioned cross-arm sensitivity for DP checkpoints trained on Concurrent data (left) and CTR data (right). Yellow circles measure right-state-to-left-Action sensitivity; blue triangles measure left-state-to-right-Action sensitivity. The two outer regions are labeled Independent and the middle region is Synchronized.}
\label{fig:phase-sensitivity-dp}
\end{figure*}

\subsection{Constraint violations and release latency}
{
For Prerequisite tasks, a violation is an attempted dependent action before the required condition is available; for Shared-Workspace Exclusion tasks, it is forbidden concurrent occupancy of the conflict region. Violation rate is the fraction of evaluations containing the task-specific violation. For a release event $e$, with $t_e^{\rm ready}$ the first time the blocked condition becomes available and $t_e^{\rm act}$ the first appropriate progress action, release latency is
\looseness=-1\par
}
\begin{equation}
 L_e=t_e^{\rm act}-t_e^{\rm ready}.
\end{equation}
Condition-level release latency is the arithmetic mean of $L_e$ over the evaluated release events in that condition.

\subsection{Paired pause effects}
For the pause intervention, a control branch and a matched branch share the same local initial state and differ only in an externally imposed pause of the opposite arm for duration $d$. For the unfrozen arm,
\begin{equation}
 \Delta T_{\rm other}(d)=T_{\rm other}^{\rm pause}(d)-T_{\rm other}^{\rm control}.
\end{equation}
{
At each duration $d\in\{1,2,3,4\}$ seconds, the reported $\Delta T_{\rm other}(d)$ is the arithmetic mean of the matched-pair differences, and $\mathrm{SR}_{\rm other}(d)$ is the unfrozen arm's local success rate at that duration. We report $\beta_{\rm delay}$ as the fitted slope of $\Delta T_{\rm other}(d)$ versus $d$ over the four durations. A value near one indicates near-complete transfer of the imposed delay; a value near zero indicates invariance. For \paperref{Figure}{fig:pause-delay-ci}, the confidence intervals are computed directly from the 50 measured delay values recorded for each condition and duration; these values are raw experimental trial records.
\looseness=-1\par
}

\subsection{Progress-state success summaries}
{
For the six-state progress-state robustness evaluation, Worst SR is the minimum of the six state-level success rates for that task and training distribution. For the held-out progress-state experiment on Sort Letters, the two arm-local progress coordinates are normalized to $[0,1]$, and we use the fixed hyperparameter choice $H=[0.20,0.30]\times[0.70,0.80]$ for the held-out region. Unseen evaluations initialize inside $H$; initial-state evaluations start from the task's standard initial state. We report success rate only for $\pi_{0.5}$ under Mixed, CTR-hole, and CTR-full.
\looseness=-1\par
}

\section{CTR implementation details}
\label{app:implementation}

\begin{algorithm}[ht]
\caption{Constraint-preserving temporal recombination}
\label{alg:ctr}
\DontPrintSemicolon
\SetKwInput{KwInput}{Input}
\SetKwInput{KwOutput}{Output}
\SetKwComment{tcp}{$\triangleright$~}{}
\SetCommentSty{textit}
\KwInput{Demonstrations $\mathcal D$, task description $\xi$, schedule law $q(\cdot\mid\tau)$, draws $J\in\mathbb N$, learner $m$}
\KwOutput{trained joint policy $\pi_{\theta_m}$}
$\widetilde{\mathcal D}\leftarrow\varnothing$\;
$\mathcal P_\xi\leftarrow\operatorname{GeneratePipeline}_{\rm GPT6}(\xi)$\tcp*[r]{once per task}
\For{$\tau\in\mathcal D$}{
  $(\Sigma_\tau,\mathcal R_{\rm req}(\tau))\leftarrow\mathcal P_\xi(\tau)$\tcp*[r]{automatic processing; no per-video GPT query}
  $\Omega(\tau)\leftarrow\operatorname{Admissible}(\Sigma_\tau,\mathcal R_{\rm req}(\tau))$\;
  \For{$j=1,\ldots,J$}{
    $s_j\leftarrow\operatorname{Sample}(q(\cdot\mid\tau))$\;
    $(\phi_L^{s_j},\phi_R^{s_j})\leftarrow\operatorname{Retime}(\tau,s_j)$\tcp*[r]{preserve local motion}
    $\widetilde\tau_j\leftarrow\mathcal C(\tau,\phi_L^{s_j},\phi_R^{s_j})$\tcp*[r]{compose real main view}
    $M_{s_j}\leftarrow\operatorname{IdleMask}(s_j,\Sigma_\tau)$\tcp*[r]{omit synthetic waits}
    $\widetilde{\mathcal D}\leftarrow\widetilde{\mathcal D}\cup\{(\widetilde\tau_j,M_{s_j})\}$\;
  }
}
$\theta_m\leftarrow\operatorname{Train}_m(\widetilde{\mathcal D};\mathcal L_m)$\;
\end{algorithm}

\paragraph{Operators and sampling.}
{
Here $\xi$ is the task description, $J$ is a finite number of schedule draws per demonstration, and $\Sigma_\tau=\{\sigma_{i,k}\}$ collects its ordered arm-local segments. GPT-6 Astra is invoked at task setup to generate the task-specific processing program $\mathcal P_\xi$; the resulting program, not GPT-6 Astra, processes each demonstration. $\mathcal P_\xi$ returns the detected segments and task-required relations $\mathcal R_{\rm req}(\tau)$; $\operatorname{Admissible}$ builds the conservative schedule set $\Omega(\tau)$ described in \paperref{Section}{sec:annotate}. Each call to $\operatorname{Sample}$ draws one $s_j$ from $q(\cdot\mid\tau)$, whose support lies in $\Omega(\tau)$; it does not enumerate a continuous schedule space. $\operatorname{Retime}$ constructs the source-time maps in \paperref{Equation}{eq:phi}; $\mathcal C$ aligns the arm-local streams and composes the real-data main view; $\operatorname{IdleMask}$ constructs $M_{s_j}$ using \paperref{Equation}{eq:idlemask}; and $\operatorname{Train}_m$ optimizes the empirical version of the model-specific objective in \paperref{Equation}{eq:objective}. In our experiments, $J\in\{1,2\}$. All conditions are trained with the matched update budget in \paperref{Appendix}{app:training-evaluation-config}; the deployed policy receives ordinary observations, and neither GPT-6 Astra nor the generated scheduler is queried online.
\looseness=-1\par
}

\subsection{GPT-6 Astra and the generated task pipeline}
{
GPT-6 Astra is used at task setup time to generate the task-specific video-processing program $\mathcal P_\xi$; it is not the per-episode video annotator. The generated program encodes the task-specific logic for interaction detection, arm-local segmentation, protected-versus-reschedulable temporal relations, recombination and scheduling, and validation. It then runs automatically on the recorded videos or datasets to produce the segment and constraint metadata consumed by CTR. Task-required cross-arm constraints include ordering, synchronized execution, sustained support, and shared-region exclusion. The resulting pipeline outputs are used only during data construction.
\looseness=-1\par
}

\subsection{Admissible schedule construction}
A schedule is rejected when it violates an annotated task-required cross-arm constraint. Relative timing between segments governed by such a constraint is conservatively held fixed. Elsewhere, segment starts may be shifted so that independent work appears in serial, overlapping, and partially overlapping bimanual progress states.

\subsection{Source-time map and recombination}
Given $s\in\Omega(\tau)$, arm $i$ is active on $\mathcal A_i(s)=\bigcup_k\Sigma_{i,k}(s)$, and the source-time map $\phi_i^s:[0,\widetilde T]\to[0,T]$ reads
\begin{equation}
 \phi_i^s(t)=
 \begin{cases}
  u_{i,1},
    & 0\le t<s_{i,1};\\[3pt]
  u_{i,k}+(t-s_{i,k}),
    & t\in\Sigma_{i,k}(s)\ \text{for some }k;\\[3pt]
  v_{i,k},
    & t\in\bigl(s_{i,k}+\ell_{i,k},\,s_{i,k+1}\bigr),\ k<K_i;\\[3pt]
  v_{i,K_i},
    & t>s_{i,K_i}+\ell_{i,K_i}.
 \end{cases}
 \label{eq:phi}
\end{equation}
On active segments, this map gives the content-preservation invariant in \paperref{Equation}{eq:preserve} of the main text.

\subsection{RobotSeg-based foreground extraction and main-view composition}
{
For each real demonstration, arm-local wrist streams remain aligned to their own source times. General-purpose segmentation and open-vocabulary detection provide relevant foreground-extraction primitives~\citep{kirillov2023sam,ravi2024sam2,liu2024groundingdino}. RobotSeg-based arm segmentation~\citep{mei2026robotseg} and the task-object annotation pipeline provide foregrounds and masks, which we select at $\phi_L^s(t)$ and $\phi_R^s(t)$ and composite over the scene background. Composition is limited to admissible schedules under the annotated task-required cross-arm constraints. Writing $F_i^s(t)=F_i(\phi_i^s(t))$ and $M_i^s(t)=M_i(\phi_i^s(t))$ for the foreground and mask of arm $i$ and scene background $B$, the recombined main view is
\looseness=-1\par
}
\begin{equation}
\begin{aligned}
 \widetilde I_t^{\rm main}
 &=\operatorname{Compose}\!\left(
 B,\ 
 M_L^s(t)\odot F_L^s(t),\ 
 M_R^s(t)\odot F_R^s(t)\right)\\
 &=\Bigl(1-\textstyle\sum_{i}M_i^s(t)\Bigr)\odot B+\sum_{i}M_i^s(t)\odot F_i^s(t),
\end{aligned}
 \label{eq:compose}
\end{equation}
the alpha-compositing form for non-overlapping masks, with occasional overlaps resolved by the annotated layering order.

\subsection{IdleMask routing}
{
IdleMask is applied only to action coordinates and timesteps whose waiting was introduced by temporal recombination. The opposite arm's valid target at the same prediction step remains supervised. Required support and post-completion holding are not masked. Formally, output time $t$ is an artificial wait for arm $i$ on an inserted pre-segment or inter-segment gap, $t\in[0,s_{i,1})\cup\bigcup_{k<K_i}\bigl(s_{i,k}+\ell_{i,k},\,s_{i,k+1}\bigr)$. For batch element $b$ with sampled schedule $s_b$, prediction step $k$ at output time $t_{bk}$, arm $i$, and action coordinate $d$,
\looseness=-1\par
}
\begin{equation}
 m_{bkid}
 =\ind\!\left[t_{bk}\in\mathcal A_i(s_b)\ \text{or}\ t_{bk}>s_{i,K_i}+\ell_{i,K_i}\right].
 \label{eq:idlemask}
\end{equation}
The masked action-supervision component is
\begin{equation}
 \mathcal L_{\rm action}
 =\frac{\sum_{b,k,i,d}m_{bkid}\,\ell_{bkid}}
        {\max(1,\sum_{b,k,i,d}m_{bkid})},
 \label{eq:loss}
\end{equation}
Here $\ell$ is the base learner's coordinatewise action, denoising, or velocity-prediction loss; model-specific auxiliary terms in $\mathcal L_m$ remain outside this mask normalization.

\subsection{Held-out progress-state construction}
{
The held-out progress-state experiment on Sort Letters uses one reschedulable block with independently normalized left- and right-arm progress. We use the fixed hyperparameter choice $H=[0.20,0.30]\times[0.70,0.80]$ for the single held-out region. No mirrored hole is used. Mixed is left unchanged and we verify that $H$ is absent from its realized bimanual progress-state support. CTR-hole explicitly removes training samples in $H$, whereas CTR-full retains full admissible support. Initial-state evaluations start from the task's standard initial state, and unseen evaluations initialize inside $H$. All three conditions use the matched real-robot $\pi_{0.5}$ training configuration of 20,000 optimizer steps described in \paperref{Appendix}{app:training-evaluation-config}; only the schedule-support construction and the stated CTR-hole filtering differ.
\looseness=-1\par
}

\section{Physical-validity and observation-consistency audit}
\label{app:physical-validity}

{
For recorded real-robot demonstrations, preserving each arm's local trajectory does not imply that an arbitrary recombination of the two trajectories is physically realizable. Likewise, a recomposed main-camera image is not identical to a concurrently captured real observation. We therefore treat physical validity and observation consistency as \emph{qualification problems}: CTR first proposes a schedule that satisfies the annotated task constraints, and the data pipeline then rejects candidates that fail geometric, contact, or observation-consistency checks. These checks make the operational schedule set deliberately conservative, but they are not a proof that every accepted trajectory would execute without deviation under closed-loop robot dynamics.
\looseness=-1\par
}

\paragraph{Conservative geometric qualification.}
{
For recorded real-robot data, we do not replay every counterfactual schedule on hardware during dataset construction. Instead, each arm advances monotonically along its recorded source clock; waits hold a recorded pose, and the scheduler does not invent arbitrary robot configurations. Newly paired left/right configurations and the motion between them are checked with the robot URDF, forward kinematics, and mesh-distance tests, together with task-specific prerequisite or shared-workspace gates. If the pipeline cannot certify the configured clearance for a newly created pairing, that schedule is rejected rather than accepted by default. These checks certify the declared geometric model and recorded-state interpolation, not unmodeled compliance, controller transients, calibration error, or contact dynamics.
\looseness=-1\par
}

\paragraph{Recorded contact phases.}
{
Some demonstrations contain intentional or unavoidable contact during a coupled manipulation. Requiring positive separation everywhere would therefore reject the original demonstration itself. We handle this case narrowly: a contact edge may be retained only when the two arm clocks reproduce the \emph{same adjacent paired source frames} from the original demonstration. Such an edge is recorded explicitly in the contact ledger. It cannot be combined with a different opposite-arm source time, skipped into, or artificially prolonged. In contrast, every newly created cross-arm edge must pass the ordinary geometric qualification. This preserves observed coupled contact without treating it as permission to synthesize new contact configurations.
\looseness=-1\par
}

\paragraph{Observation composition and object state.}
{
Only the shared main view is spatially recomposed. The compositor maintains an explicit ownership model for task objects: an object is scene content before pickup, joins the corresponding arm foreground while it is carried, and returns to shared-scene ownership after release. The pipeline also reconstructs the object's source region after pickup and checks for duplicated objects at their original location. Where registered metric depth is available, overlapping arm foregrounds use depth for per-pixel occlusion ordering; otherwise the fallback ordering is recorded explicitly. Candidates can therefore be rejected even when timing is geometrically admissible---for example, when no safe schedule is certified or when the main-view composition audit indicates an implausible duplicate or otherwise unreliable composition.
\looseness=-1\par
}

{
The wrist cameras have a narrower guarantee. Each wrist stream follows its own arm's recorded source clock and is checked against the corresponding source video, but we currently do \emph{not} recomposite the opposite arm when it happens to be visible in that wrist camera. Consequently, wrist-local content is source-consistent, while cross-arm appearance inside a wrist image can differ from the fully realized counterfactual scene. We treat this as a known approximation rather than as a physically exact observation model.
\looseness=-1\par
}

\end{document}